\documentclass[manuscript,nonacm]{acmart}

\usepackage{xcolor}
\usepackage{appendix}

\usepackage{makecell}

\usepackage{amsmath}
\usepackage{subfig}
\usepackage{xfrac} 
\usepackage{tablefootnote}
\usepackage{multirow}
\usepackage{xspace}

\DeclareMathOperator*{\argmax}{argmax}

\newcommand{\mypar}[1]{\vskip2mm\noindent{\bf #1}\xspace}

\usepackage{xspace}
\def\RESCAL{RESCAL\xspace}
\def\TRANSE{TransE\xspace}

\begin{document}

\title{Knowledge Graph Embedding for Link Prediction: A Comparative Analysis}

\author{Andrea Rossi}
\affiliation{\institution{Roma Tre University}
}
\email{andrea.rossi3@uniroma3.it}

\author{Donatella Firmani}
\affiliation{\institution{Roma Tre University}
}
\email{donatella.firmani@uniroma3.it}

\author{Antonio Matinata}
\affiliation{\institution{Roma Tre University}
}
\email{ant.matinata@stud.uniroma3.it}

\author{Paolo Merialdo}
\affiliation{\institution{Roma Tre University}
}
\email{paolo.merialdo@uniroma3.it}

\author{Denilson Barbosa}
\affiliation{\institution{University of Alberta}
}
\email{denilson@ualberta.ca}

\renewcommand{\shortauthors}{Rossi, et al.}

\begin{abstract}
Knowledge Graphs (KGs) have found many applications in industry and academic settings, which in turn, have motivated considerable research efforts towards large-scale information extraction from a variety of sources. Despite such efforts, it is well known that even state-of-the-art KGs suffer from incompleteness. Link Prediction (LP), the task of predicting missing facts among entities already a KG, is a promising and widely studied task aimed at addressing KG incompleteness. 
Among the recent LP techniques, those based on \emph{KG embeddings} have achieved very promising performances in some benchmarks. Despite the fast growing literature in the subject, insufficient attention has been paid to the effect of the various design choices in those methods. 
Moreover, the standard practice in this area is to report accuracy by aggregating over a large number of test facts in which some entities are over-represented; this allows LP methods to exhibit good performance by just attending to structural properties that include such entities, while ignoring the remaining majority of the KG.
This analysis provides a comprehensive comparison of embedding-based LP methods, extending the dimensions of analysis beyond what is commonly available in the literature. 
We experimentally compare effectiveness and efficiency of 16 state-of-the-art methods, consider a rule-based baseline, and report detailed analysis over the most popular benchmarks in the literature. \end{abstract}


\maketitle

\section{Introduction}
\label{sec:intro}

Knowledge Graphs (KGs) are structured representations of real world information. In a KG \emph{nodes} represent entities, such as people and places; \emph{labels} are types of relations that can connect them; \emph{edges} are specific facts connecting two entities with a relation. 
Due to their capability to model structured, complex data in a machine-readable way, KGs are nowadays widely employed in various domains, ranging from question answering to information retrieval and content-based recommendation systems, and they are vital to any semantic web project~\cite{Hovy_2013_CBS_2405838_2405907}. Examples of notable KGs are FreeBase~\cite{freebase}, WikiData~\cite{wikidata}, DBPedia~\cite{dbpedia}, Yago~\cite{yago} and -- in industry -- Google KG~\cite{googlekg}, Microsoft Satori~\cite{satori} and Facebook Graph Search~\cite{facebookkg}. These massive KGs can contain millions of entities and billions of facts.

Despite such efforts, it is well known that even state-of-the-art KGs suffer from \emph{incompleteness}. For instance, it has been observed that over 70\% of person entities have no known place of birth, and over 99\% have no known ethnicity~\cite{incompleteness,dong2014} in FreeBase, one of the largest and most widely used KGs for research purposes. This has led researchers to propose various techniques for correcting errors as well as adding missing facts to KGs~\cite{paulheim2017knowledge}, commonly known as the task of \emph{Knowledge Graph Completion} or \emph{Knowledge Graph Augmentation}.
Growing an existing KG can be done by extracting new facts from external sources, such as Web corpora, or by inferring missing facts from those already in the KG. The latter approach, called \emph{Link Prediction} (LP), is the focus of our analysis.

LP has been an increasingly active area of research, which has more recently benefited from the explosion of machine learning and deep learning techniques. The vast majority of LP models nowadays use original KG elements to learn low-dimensional representations dubbed \emph{Knowledge Graph Embeddings}, and then employ them to infer new facts.
Inspired by a few seminal works such as \RESCAL~\cite{rescal} and \TRANSE~\cite{transe}, in the short span of just a few years researchers have developed dozens of novel models based on very different architectures.
One aspect that is common to the vast majority of papers in this area, but nevertheless also problematic, is that they report results  aggregated over a large number of test facts in which few entities are over-represented. 
As a result, LP methods can exhibit good performance on these benchmarks by attending only to such entities while ignoring the others.
Moreover, the limitations of the current best-practice can make it difficult for one to understand how the papers in this literature fit together and to picture what research directions are worth pursuing.
In addition to that, the strengths, weaknesses and limitations of the current techniques are still unknown, that is, the circumstances allowing models to perform better have been hardly investigated. Roughly speaking, we still do not really know what makes a fact easy or hard to learn and predict.

In order to mitigate the issues mentioned above, we carry out an extensive comparative analysis of a representative set of LP models based on KG embeddings. We privilege state-of-the-art systems, and consider works belonging to a wide range of architectures. We train and tune such systems from scratch and provide experimental results beyond what is available in the original papers, by proposing new and informative evaluation practices. Specifically:
\begin{itemize}
    \item  We take into account 16 models belonging to diverse machine learning and deep learning architectures; we also adopt as a baseline an additional state-of-the-art LP model based on rule mining. We provide a detailed description of the approaches considered for experimental comparison and a summary of related literature, together with an educational taxonomy for Knowledge Graph Embedding techniques.
    \item We take into account the 5 most commonly employed datasets as well as the most popular metrics currently used for benchmarking; we analyze in detail their features and peculiarities.
    \item For each model we provide quantitative results for efficiency and effectiveness on every dataset.
    \item We define a set of structural features in the training data, and we measure how they affect the predictive performance of each model on each test fact. \end{itemize}

The datasets, the code and all the resources used in our work are publicly available through our GitHub repository.\footnote{https://github.com/merialdo/research.lpca. For each model and dataset, we also share CSV files containing, for each test prediction, the rank and the list of all the entities predicted up to the correct one.}

\paragraph{Outline} The paper is organized as follow. Section~\ref{sec:problem_overview} provides background on KG embedding and LP. Section~\ref{sec:taxonomy} introduces the models included in our work, presenting them in a taxonomy to facilitate their description. Section~\ref{sec:methodology} describes the analysis directions and approaches we follow in our work. Section~\ref{sec:experiments} reports our results and observations.
Section~\ref{sec:lessons} provides lessons learned and future research directions. Section~\ref{sec:related} discusses related works, and Section~\ref{sec:concl} provides concluding remarks.

\section{The Link Prediction Problem}
\label{sec:problem_overview}
This section provides a detailed outline for the LP task in the context of KGs, introducing key concepts that we are going to refer to in our work.

We define a KG as a labeled, directed multi-graph $KG = (\mathcal{E}, \mathcal{R}, \mathcal{G})$:
\begin{itemize}
    \item $\mathcal{E}$: a set of nodes representing \emph{entities};
    \item $\mathcal{R}$: a set of labels representing \emph{relations};
    \item $\mathcal{G} \subseteq \mathcal{E} \times \mathcal{R} \times \mathcal{E}$: a set of edges representing \emph{facts} connecting pairs of entities. Each fact is a triple \textlangle{}\textit{h}, \textit{r}, \textit{t}\textrangle{}, where \textit{h} is the \emph{head}, \textit{r} is the \emph{relation}, and \textit{t} is the \emph{tail} of the fact.
\end{itemize}

\textbf{Link Prediction} (LP) is the task of exploiting the existing facts in a KG to infer missing ones. This amounts to guessing the correct entity that completes \textlangle{}\textit{h}, \textit{r}, ?\textrangle{} (tail prediction) or \textlangle{}?, \textit{r}, \textit{t}\textrangle{} (head prediction).
For the sake of simplicity, when talking about head and tail prediction globally, we call \emph{source} entity the known entity in the prediction, and \emph{target} entity the one to predict. 

In time, numerous approaches have been proposed to tackle the LP task. Some methods are based on observable features and employ techniques such as Rule Mining~\cite{amie2013}\cite{amieplus2015}\cite{rulen2018}\cite{huynh2019mining} or the Path Ranking Algorithm~\cite{pra2010}\cite{pra2011} to identify missing triples in the graph. 
Recently, with the rise of novel Machine Learning techniques, researchers have been experimenting on capturing latent features of the graph with vectorized representations, or embeddings, of its components. 
In general, \emph{embeddings} are vectors of numerical values that can be used to represent any kind of elements (e.g., depending on the domain: words, people, products...). Embeddings are learned automatically, based on how the corresponding elements occur and interact with each other in datasets representative of the real world. For instance, word embeddings have become a standard way to represent words in a vocabulary, and they are usually learned using textual corpora as input data. When it comes to KGs, embeddings are typically used to represent entities and relationships using the graph structure; the resulting vectors, dubbed \textbf{KG Embeddings}, embody the semantics of the original graph, and can be used to identify new links inside it, thus tackling the LP task.

In the following we use $italic$ letters to identify KG elements (entities or relations), and $\boldsymbol{bold}$ letters to identify the corresponding embeddings. Given for instance a generic entity, we may use $\textit{e}$ when referring to its element in the graph, and $\boldsymbol{e}$ when referring to its embedding.

Datasets employed in LP research are typically obtained subsampling real-world KGs; each dataset can therefore be seen as a small KG with its own sets of entities $\mathcal{E}$, relations $\mathcal{R}$ and facts $\mathcal{G}$. In order to facilitate research, $\mathcal{G}$ is further split into three disjoint subsets: a training set $\mathcal{G}_{train}$, a validation set $\mathcal{G}_{valid}$ and a test set $\mathcal{G}_{test}$.

Most of LP models based on embeddings define a scoring function $\phi$ to estimate the plausibility of any fact  $\langle\textit{h},~\textit{r},~\textit{t}\rangle$ using their embeddings:
\[\phi(\boldsymbol{h}, \boldsymbol{r}, \boldsymbol{t})\]
In this paper, unless differently specified, we are going to assume that the higher the score of $\phi$, the more plausible the fact. 

During training, embeddings are usually initialized randomly and subsequently improved with optimization algorithms such as back-propagation with gradient descent. The positive samples in $\mathcal{G}_{train}$ are often randomly corrupted in order to generate negative samples. The optimization process aims at maximizing the plausibility of positive facts as well as minimizing the plausibility of negative facts; this often amounts to employing a \textbf{triplet loss} function. Over time, more effective ways to generate negative triples have been proposed, such as sampling from a Bernouilli distribution~\cite{transh} or generating them with adversarial algorithms~\cite{rotate}.
In addition to the embeddings of KG elements, models may also use the same optimization algorithms to learn additional parameters (e.g. the weights of neural layers). Such parameters, if present, are employed in the scoring function $\phi$ to process the actual embeddings of entities and relations. Since they are not specific to any KG element, they are often dubbed \emph{shared parameters}. 

In prediction phase, given an incomplete triple $\langle\textit{h}, \textit{r}, ?\rangle$, the missing tail is inferred as the entity that, completing the triple, results in the highest score:
\[t = \argmax_{e \in \mathcal{E}} \phi(\boldsymbol{h}, \boldsymbol{r}, \boldsymbol{e})\]
Head prediction is performed analogously.

Evaluation is carried out by performing both head and tail prediction on all test triples in $\mathcal{G}_{test}$, and computing for each prediction how the target entity ranks against all the other ones. Ideally, the target entity should yield the highest plausibility.

Ranks can be computed in two largely different settings, called raw and filtered scenarios. As a matter of fact, a prediction may have multiple valid answers: for instance, when predicting the tail for \textlangle{}~\emph{Barack~Obama},~\emph{parent},~\emph{Natasha~Obama}~\textrangle{}, a model may associate a higher score to \emph{Malia~Obama} than to \emph{Natasha~Obama}.
More generally, if the predicted fact is contained in $\mathcal{G}$ (that is, either in $\mathcal{G}_{train}$, or in $\mathcal{G}_{valid}$ or in $\mathcal{G}_{test}$), the answer is valid. Depending on whether valid answers should be considered acceptable or not, two separate settings have been devised:
\begin{itemize}
    \item \emph{Raw Scenario}: in this scenario, valid entities outscoring the target one are considered as mistakes. Therefore they do contribute to the rank computation. Given a test fact \textlangle{}$h,~r,~t$\textrangle{}$~\in~\mathcal{G}_{test}$, the raw rank $r_{t}$ of the target tail $t$ is computed as:
    \[r_{t} = |\{e\in\mathcal{E}\setminus\{t\}~:~ \phi(\boldsymbol{h},\boldsymbol{r},\boldsymbol{e}) > \phi(\boldsymbol{h},\boldsymbol{r},\boldsymbol{t}) \}| + 1\]
    The raw rank in head prediction can be computed analogously.
    
    \item \emph{Filtered Scenario}: in this scenario, valid entities outscoring the target one are not considered mistakes. Therefore they are skipped when computing the rank. Given a test fact \textlangle{}$h,~r,~t$\textrangle{}$~\in~\mathcal{G}_{test}$, the filtered rank $r_{t}$ of the target tail $t$ is computed as:
    \[r_{t} = |\{e\in\mathcal{E}\setminus\{t\}~:~\phi(\boldsymbol{h},\boldsymbol{r},\boldsymbol{e}) > \phi(\boldsymbol{h},\boldsymbol{r},\boldsymbol{t})~\wedge~\langle~h,~r,~e~\rangle\notin\mathcal{G}\}|+1\]
    The filtered rank in head prediction can be computed analogously.
\end{itemize}

In order to compute the rank it is also necessary to define the policy to apply when the target entity obtains the same score as other ones. This event is called a \emph{tie} and it can be handled with different policies:
\begin{itemize}
    \item \emph{min}: the target is given the lowest rank among the entities in tie. This is the most permissive policy, and it may result in artificially boosting performances: as an extreme example, a model systematically setting the same score to all entities would obtain perfect results under this policy.
    \item \emph{average}: the target is given the average rank among the entities in tie. 
    \item \emph{random}: the target is given a random rank among the entities in tie. On large test sets, this policy should globally amount to the average policy.
    \item \emph{ordinal}: the entities in tie are given ranks based on the order in which they have been passed to the model. This usually depends on the internal identifiers of entities, which are independent from their scores: therefore this policy should globally correspond to the random policy.
    \item \emph{max}: the target is given the highest (worst) rank among the entities in tie. This is the most strict policy.
\end{itemize}

The ranks $Q$ obtained from test predictions are usually employed to compute standard global metrics. The most commonly employed metrics in LP are:

\paragraph{Mean Rank (MR)} 
It is the average of the obtained ranks: 
\[MR = \frac{1}{|Q|} \sum_{q \in Q} q\] 
It is always between 1 and $|\mathcal{E}|$, and the lower it is, the better the model results. It is very sensitive to outliers, therefore researchers lately have started avoiding it, resorting to Mean Reciprocal Rank instead.

\paragraph{Mean Reciprocal Rank (MRR)}
It is the average of the inverse of the obtained ranks: 
\[MRR = \frac{1}{|Q|} \sum_{q \in Q} \frac{1}{q}\] 
It is always between 0 and 1, and the higher it is, the better the model results.

\paragraph{Hits@K (H@K)}
It is the ratio of predictions for which the rank is equal or lesser than a threshold $K$: 
\[H@K~=~\frac{|\{q~\in~Q~:~q~\leq~K\}|}{|Q|}\] 
Common values for K are $1, 3, 5, 10$. 
The higher the H@K, the better the model results.
In particular, when $K~=~1$, it measures the ratio of the test facts in which the target was predicted correctly on the first try. H@1 and MRR are often closely related, because these predictions also correspond to the most relevant addends to the MRR formula.

These metrics can be computed either separately for subsets of predictions (e.g. considering separately head and tail predictions) or considering all test predictions altogether.

\section{Overview of Link Prediction Techniques}
\label{sec:taxonomy}

In this section we survey and discuss the main LP approaches for KGs based on latent features. As already described in Section~\ref{sec:problem_overview}, LP models can exploit a large variety of approaches and architectures, depending on how they model the optimization problem and on the techniques they implement to tackle it. 

\begin{figure}
    \includegraphics[width=\textwidth]{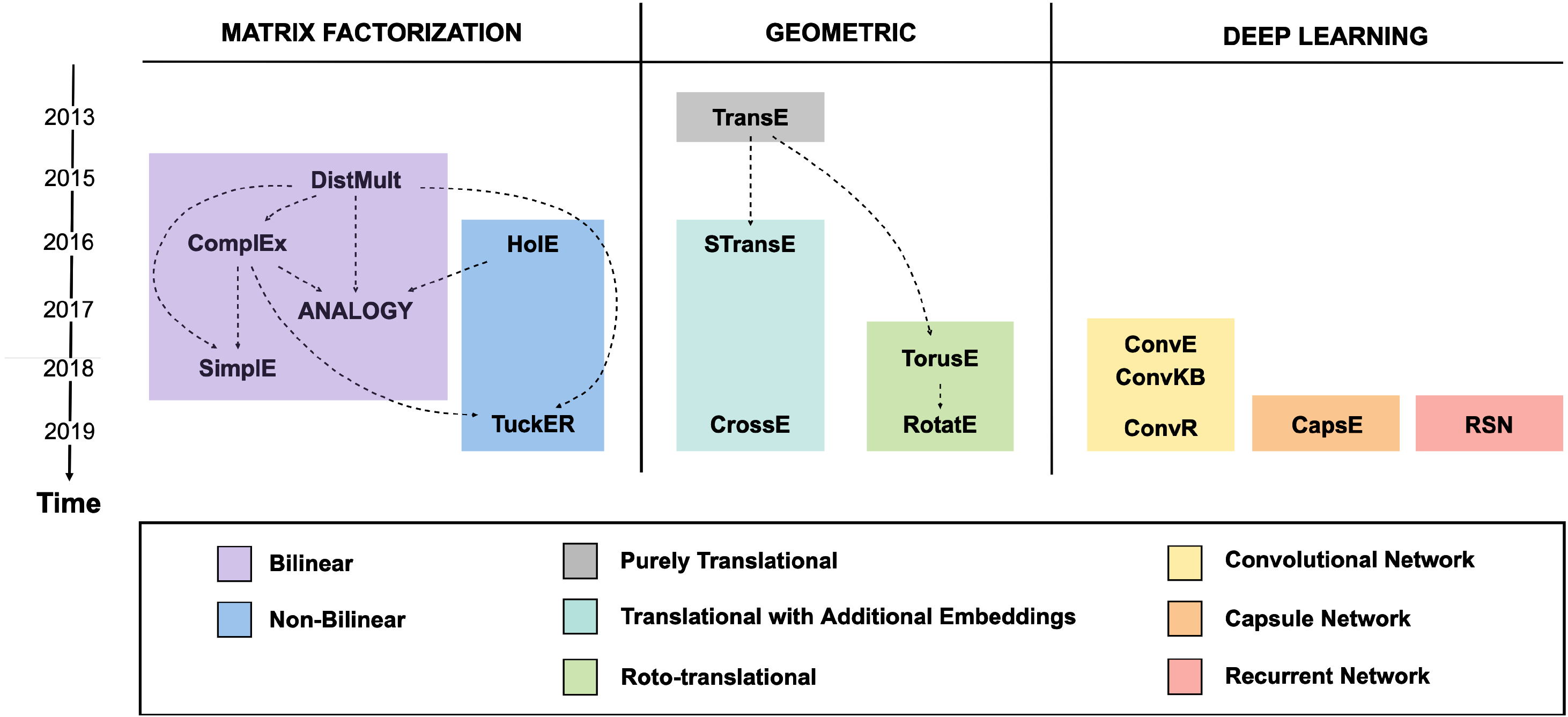}
    \caption{\label{fig:taxonomy} Taxonomy for the LP models included in our analysis. Dotted arrows indicate that the target method builds on the source method by either generalizing or specializing the definition of its scoring function. The included models are: DistMult~\protect\cite{distmult}; 
    ComplEx~\protect\cite{complex}; 
    ANALOGY~\protect\cite{analogy}; 
    SimplE~\protect\cite{simple};
    HolE~\protect\cite{hole};
    TuckER~\protect\cite{tucker};
    TransE~\protect\cite{transe};
    STransE~\protect\cite{stranse};
    CrossE~\protect\cite{crosse};
    TorusE~\protect\cite{toruse};
    RotatE~\protect\cite{rotate};
    ConvE~\protect\cite{dettmers2018convolutional};
    ConvKB~\protect\cite{convkb};
    ConvR~\protect\cite{convr};
    CapsE~\protect\cite{capse};
    RSN~\protect\cite{rsn}.}
\end{figure}

In order to overview their highly diverse characteristics we propose a novel taxonomy illustrated in Figure~\ref{fig:taxonomy}. We define three main families of models, and further divide each of them into smaller groups, identified by unique colours. For each group, we include the most valid representative models, prioritizing the ones reaching state-of-the-art performance and, whenever possible, those with publicly available implementations. The result is a set of 16 models based on extremely diverse architectures; these are the models we subsequently employ in the experimental sections of our comparative analysis. For each model we also report the year of publication as well as the influences it has received from the others. We believe that this taxonomy facilitates the understanding of these models and of the experiments carried out in our work.

Further information on the included models, such as their loss function and their space complexity, is reported in Table~\ref{tab:loss_table}.

In our analysis we focus on the body of literature for systems that learn from the KG structure. We refer the reader to works discussing how to leverage additional sources of information, such as textual captions~\cite{toutanova_jointly},\cite{tekeh},\cite{atec}, images~\cite{ikrl} or pre-computed rules~\cite{ruge}; see~\cite{gesese} for a survey exclusive to these models. 

We identify three main families of models: 1) \emph{Tensor Decomposition Models}; 2) \emph{Geometric Models}; 3) \emph{Deep Learning Models}.

\subsection{Tensor Decomposition Models} 
Models in this family interpret LP as a task of tensor decomposition~\cite{tensordecomposition}. 
These models implicitly consider the KG as a 3D adjacency matrix (that is, a 3-way tensor), that is only only partially observable due to the KG incompleteness. The tensor is decomposed into a combination (e.g. a multi-linear product) of low-dimensional vectors: such vectors are used as embeddings for entities and relations. The core idea is that, provided that the model does not overfit on the training set, the learned embeddings should be able to generalize, and associate high values to unseen true facts in the graph adjacency matrix. In practice, the score of each fact is computed operating that combination on the specific embeddings involved in that fact; the embeddings are learned as usual by optimizing the scoring function for all training facts. These models tend to employ few or no shared parameters at all; this makes them particularly light and easy to train.

\subsubsection{Bilinear Models} Given the head embedding $\boldsymbol{h} \in \mathbb{R}^d$ and the tail embedding $\boldsymbol{t} \in \mathbb{R}^d$, these models represent the relation embedding as a bidimensional matrix $\boldsymbol{r} \in \mathbb{R}^{d \times d}$. The scoring function is then computed as a bilinear product:
\[\phi(\boldsymbol{h}, \boldsymbol{r}, \boldsymbol{t}) = \boldsymbol{h} \times \boldsymbol{r} \times \boldsymbol{t} \]
where symbol $\times$ denotes matrix product. 
These models usually differ from one another by introducing specific additional constraints on the embeddings they learn.
For this group, in our comparative analysis, we include the following representative models:

\mypar{DistMult}~\cite{distmult} forces all relation embeddings to be diagonal matrices, which consistently reduces the space of parameters to be learned, resulting in a much easier model to train. On the other hand, this makes the scoring function commutative, with $\phi(\boldsymbol{h}, \boldsymbol{r}, \boldsymbol{t}) = \phi(\boldsymbol{t}, \boldsymbol{r}, \boldsymbol{s})$, which amounts to treating all relations as symmetric. Despite this flaw, it has been demonstrated by Kadlec~\emph{et~al}.~\cite{kadlec2017knowledge} that, when carefully tuned, DistMult can still reach state-of-the-art performance.
        
\mypar{ComplEx}~\cite{complex}, similarly to DistMult, forces each relation embedding to be a diagonal matrix, but extends such formulation in the complex space: $\boldsymbol{h} \in \mathbb{C}^d$, $\boldsymbol{t} \in \mathbb{C}^d$, $\boldsymbol{r} \in \mathbb{C}^{d \times d}$.
In the complex space, the bilinear product becomes a Hermitian product, where in lieu of the traditional $\boldsymbol{t}$, its conjugate-transpose $\boldsymbol{\overline{t}}$ is employed. This disables the commutativeness above mentioned for the scoring function, allowing ComplEx to successfully model asymmetric relations as well.

\mypar{Analogy}~\cite{analogy} aims at modeling analogical reasoning, which is key for any kind of knowledge induction. It employs the general bilinear scoring function but adds two main constraints inspired by analogical structures:
$\boldsymbol{r}$ must be a normal matrix: $\boldsymbol{r}\boldsymbol{r}^T = \boldsymbol{r}^T\boldsymbol{r}$; for each pair of relations $r_1$, $r_2$, their composition must be commutative: $\boldsymbol{r_1} \circ \boldsymbol{r_2} = \boldsymbol{r_2} \circ \boldsymbol{r_1}$. The authors demonstrate that normal matrices can be successfully employed for modelling asymmetric relations.

\mypar{SimplE}~\cite{simple} forces relation embeddings to be diagonal matrices, similarly to DistMult, but extends it by $(i)$
associating with each entity $e$ two separate embeddings, $\boldsymbol{e_h}$ and $\boldsymbol{e_t}$, depending on whether $e$ is used as head or tail; $(ii)$
associating with each relation $r$ two separate diagonal matrices, $\boldsymbol{r}$ and $\boldsymbol{r_{-1}}$, expressing the relation in its regular and inverse direction.
The score of a fact is computed averaging the bilinear scores of the regular fact and its inverse version. It has been demonstrated that SimplE is fully expressive, and therefore, unlike DistMult, it can model also asymmetric relations.  

\subsubsection{Non-bilinear Models}
These models combine the head, relation and tail embeddings of composition using formulations different from the strictly bilinear product.

\mypar{HolE}~\cite{hole}, instead of using bilinear products, computes circular correlation (denoted by $\star$ in Table~\ref{tab:loss_table}) between the embeddings of head and tail entities; then, it performs matrix multiplication with the relation embedding. Note that in this model the relation embeddings have the same shape as the entity embedding.
The authors point out that circular correlation can be seen as a compression of the full matrix product: this makes HolE less expensive than an unconstrained bilinear model in terms of both time and space complexity.

\mypar{TuckER}~\cite{tucker} relies on the Tucker decomposition~\cite{tuckerdecomposition}, which factorizes a tensor into a set of vectors and a smaller shared core $\boldsymbol{W}$.
The TuckER model learns $\boldsymbol{W}$ jointly with the KG embeddings. As a matter of fact, learning globally shared parameters is rather uncommon in Matrix Factorization Models; the authors explain that $\boldsymbol{W}$ can be seen as a shared pool of prototype relation matrices, that get combined in a different way for each relation depending in its embedding.
In TuckER the dimensions of entity and relation embeddings are independent from each other, with entity embeddings $\boldsymbol{e} \in \mathbb{R}^{d_e}$ and relation embeddings $\boldsymbol{r} \in \mathbb{R}^{d_r}$. The shape of $\boldsymbol{W}$ depends on the dimensions of entities and relations, with $\boldsymbol{W} \in \mathbb{R}^{d_e \times d_r \times d_e}$. In Table~\ref{tab:loss_table}, we denote with $\times_{i}$ the tensor product along mode $i$ used by TuckER.

\subsection{Geometric Models} 
Geometric Models interpret relations as geometric transformations in the latent space. Given a fact, the head embedding undergoes a spatial transformation $\tau$ that uses the values of the relation embedding as parameters. The fact score is the distance between the resulting vector and the tail vector; such an offset is computed using a distance function $\delta$ (e.g. L1 of L2 norm).
\[\phi(\boldsymbol{h}, \boldsymbol{r}, \boldsymbol{t}) = \delta(\tau(\boldsymbol{h}, \boldsymbol{r}), \boldsymbol{t}) \]
		
Depending on the analytical form of $\tau$, Geometric models may share similarities with Tensor Decomposition models, but in these cases geometric models usually need to enforce additional constraints in order to make their $\tau$ implement a valid spatial transformation. For instance, the rotation operated by model RotatE can be formulated as a matrix product, but the rotation matrix would need to be diagonal and to have elements with modulus 1.

Much like with Matrix Factorization Models, these systems usually avoid shared parameters, running back-propagation directly on the embeddings.
We identify three groups in this family:   \emph{(i)} \emph{Pure Translational Models},  \emph{(ii)} \emph{Translational Models with Additional Embeddings}, and     \emph{(iii)} \emph{Roto-translational models}.

\subsubsection{Pure Translational Models} 
These models interpret each relation as a \textit{translation} in the latent space: the relation embedding is just added to the head embedding, and we expect to land in a position close to the tail embedding. 
These models thus represent entities and relations as one-dimensional vectors of same length.

\mypar{TransE}~\cite{transe} was the first LP model to propose a geometric interpretation of the latent space, largely inspired by the capability observed in Word2vec vectors~\cite{word2vec} to capture relations between words in the form of translations between their embeddings. TransE enforces this explicitly, requiring that the tail embedding lies close to the sum of the head and relation embeddings, according to the chosen distance function.
Due to the nature of translation, TransE is not able to correctly handle one-to-many and many-to-one relations, as well as symmetric and transitive relations.

\subsubsection{Translational models with Additional Embeddings} These models may associate more than one embedding to each KG element. This often amounts to using specialized embeddings, such as relation-specific embeddings for each entity or, vice-versa, entity-specific embeddings for each relation. As a consequence, these models overcome the limitations of purely translational models at the cost of learning a larger number of parameters.

\mypar{STransE}~\cite{stranse}, in addition to the $d$-sized embeddings seen in TransE, associates to each relation $r$ two additional $d \times d$ independent matrices $\boldsymbol{W_{r}^{h}}$ and $\boldsymbol{W_{r}^{t}}$. When computing the score of a fact $\langle h, r, t \rangle$, before operating the usual translation, $\boldsymbol{h}$ is pre-multiplied by $\boldsymbol{W_{r}^{h}}$ and $\boldsymbol{t}$ by $\boldsymbol{W_{r}^{t}}$; this amounts to use relation-specific embeddings for the head and tail, alleviating the issues suffered by TransE on 1-to-many, many-to-one and many-to-many relations. 

\mypar{CrossE}~\cite{crosse} is one of the most recent and also most effective models in this group. For each relation it learns an additional relation-specific embedding $\boldsymbol{c_r}$. Given any fact $\langle h, r, t \rangle$, CrossE uses element-wise products (denoted by $\odot$ in Table~\ref{tab:loss_table}) to combine $\boldsymbol{h}$ and $\boldsymbol{r}$ with $\boldsymbol{c_r}$. This results in triple-specific embeddings, dubbed interaction embeddings, that are then used in the translation. Interestingly, despite not relying on neural layers, this model adopts the common deep learning practice to interpose operations with non-linear activation functions, such as \textit{hyperbolic tangent} and \textit{sigmoid} denoted (denoted respectively by $\tanh$ and $\sigma$ in Table~\ref{tab:loss_table}).

\subsubsection{Roto-Translational Models}
These models include operations that are not directly expressible as pure translations: this often amounts to perform rotation-like transformations either in combination or in alternative to translations.

\mypar{TorusE}~\cite{toruse} was motivated by the observation that the regularization used in TransE forces entity embeddings to lie on a hypersphere, thus limiting their capability to satisfy the translational constraint.
To solve this problem, TorusE projects each point $\boldsymbol{x}$ of the traditional open manifold $\mathbb{R}^d$ into a $\boldsymbol{[x]}$ point on a torus $\mathbb{T}^d$.
The authors define torus distance functions $d_{L1}$, $d_{L2}$ and $d_{eL2}$, corresponding to L1, L2 and squared L2 norm respectively (we report in Table~\ref{tab:loss_table} the scoring function with the extended form of $d_{L1}$). 
        
\mypar{RotatE}~\cite{rotate} represents relations as rotations in a complex latent space, with $\boldsymbol{h}$, $\boldsymbol{r}$ and  $\boldsymbol{t}$ all belonging to $\mathbb{C}^d$. 
The $\boldsymbol{r}$ embedding is a rotation vector: in all its elements, the complex component conveys the rotation along that axis, whereas the real component is always equal to 1. The rotation $\boldsymbol{r}$ is applied to $\boldsymbol{h}$ by operating an element-wise product (once again noted with $\odot$ in \ref{tab:loss_table}). L1 norm is used for measuring the distance from $\boldsymbol{t}$. The authors demonstrate that rotation allows to model correctly numerous relational patterns, such as symmetry/anti-symmetry, inversion and composition.

\subsection{Deep Learning Models} 

Deep Learning Models use deep neural networks to perform the LP task. 
Neural Networks learn parameters such as weights and biases, that they combine with the input data in order to recognize significant patterns. Deep neural networks usually organize parameters into separate layers, generally interspersed with non-linear activation functions. 

In time, numerous types of layers have been developed, applying very different operations to the input data. 
Dense layers, for instance, will just combine the input data $\boldsymbol{X}$ with weights $\boldsymbol{W}$ and add a bias $\boldsymbol{B}$: $\boldsymbol{W}\times\boldsymbol{X}+\boldsymbol{B}$. For the sake of simplicity, in the following formulas we will not mention the use of bias, keeping it implicit. 
More advanced layers perform more complex operations, such as convolutional layers, that learn convolution kernels to apply to the input data, or recurrent layers, that handle sequential inputs in a recursive fashion.

In the LP field, KG embeddings are usually learned jointly with the weights and biases of the layers; these shared parameters make these models more expressive, but potentially heavier, harder to train, and more prone to overfitting. We identify three groups in this family, based on the neural architecture they employ: \emph{(i)} \emph{Convolutional Neural Networks}, \emph{(ii)} \emph{Capsule Neural Networks}, and \emph{(iii)} \emph{Recurrent Neural Networks}.

\subsubsection{Convolutional Neural Networks} These models use one or multiple convolutional layers~\cite{convnets}: each of these layers performs convolution on the input data (e.g. the embeddings of the KG elements in a training fact) applying low-dimensional filters $\boldsymbol{\omega}$. The result is a \textit{feature map} that is usually then passed to additional dense layers in order to compute the fact score.

\mypar{ConvE}~\cite{dettmers2018convolutional} represents entities and relations as one-dimensional $d$-sized embeddings. When computing the score of a fact, it concatenates and reshapes the head and relation embeddings $\boldsymbol{h}$ and $\boldsymbol{r}$ into a unique input $[\boldsymbol{h}; \boldsymbol{r}]$; we dub the resulting dimensions $d_m \times d_n$. 
This input is let through a convolutional layer with a set $\boldsymbol{\omega}$ of $m \times n$ filters, and then through a dense layer with $d$ neurons and a set of weights $W$. The output is finally combined with the tail embedding $\boldsymbol{t}$ using dot product, resulting in the fact score. When using the entire matrix of entity embeddings instead of the embedding of just the one target entity $\boldsymbol{t}$, this architecture can be seen as a classifier with $|\mathcal{E}|$ classes. 

\mypar{ConvKB}~\cite{convkb} models entities and relations as same-sized one-dimensional embeddings. Differently from ConvE, given any fact $\langle h, r, t \rangle$, it concatenates all their embeddings $\boldsymbol{h}$, $\boldsymbol{r}$ and $\boldsymbol{t}$ into a $d \times 3$ input matrix $[\boldsymbol{h};\boldsymbol{r};\boldsymbol{t}]$.
This input is passed to a convolutional layer with a set $\boldsymbol{\omega}$ of $T$ filters of shape $1 \times 3$, resulting in a $T \times 3$ feature map. The feature map is let through a dense layer with only one neuron and weights $\boldsymbol{W}$, resulting in the fact score. This architecture can be seen as a binary classifier, yielding the probability that the input fact is valid.
        
\mypar{ConvR}~\cite{convr} represents entity and relation embeddings as one-dimensional vectors of different dimensions $d_e$ and $d_r$. For any fact $\langle h, r, t \rangle$, $\boldsymbol{h}$ is first reshaped into a matrix of shape $d_{e_m}, d_{e_n}$, where $d_{e_m} \times d_{e_n} = d_e$. 
$\boldsymbol{r}$ is then reshaped and split into a set $\boldsymbol{\omega_{r}}$ of $T$ convolutional filters, each of which has size $m \times n$. These filters are then employed to run convolution on $\boldsymbol{h}$; this amounts to performing an adaptive convolution with relation-specific filters.
The resulting feature maps are passed to a dense layer with weights $W$, As in ConvE, the fact score is obtained combining the neural output with the tail embedding $\boldsymbol{t}$ using dot product. 

\subsubsection{Capsule Neural Networks} Capsule networks (CapsNets) are composed of groups of neurons, called \textbf{capsules}, that encode specific features of the input, such as the presence of a specific object in an image~\cite{capsnets}. CapsNets are designed to recognize such features without losing spatial information the way that convolutional networks do. Each capsule sends its output to higher order ones, with connections decided by a dynamic routing process. The probability of a capsule detecting the feature is given by the length of its output vector.

\mypar{CapsE}~\cite{capse} embeds entities and relations into $d$-sized one-dimensional vectors, under the basic assumption that different embeddings encode homologous aspects in the same positions. Similarly to ConvKB, it concatenates $\boldsymbol{h}$, $\boldsymbol{r}$ and $\boldsymbol{t}$ into one $d \times 3$ input matrix.
This is let through a convolutional layer with $E$ $1 \times 3$ filters. The result is a $d \times E$ matrix in which the $i$-th value of any row uniquely depends on $\boldsymbol{h}[i]$, $\boldsymbol{r}[i]$ and $\boldsymbol{t}[i]$.
The matrix is let through a capsule layer; a separate capsule handles each column, thus receiving information regarding one aspect of the input fact. A second layer with one capsule is used to yield the triple score. In Table~\ref{tab:loss_table}, we denote the capsule layers with $capsnet$.

\subsubsection{Recurrent Neural Networks (RNNs)} These models employ one or multiple recurrent layers~\cite{recurrent} to analyze entire paths (sequences of facts) extracted from the training set, instead of just processing individual facts separately.

\mypar{RSN}~\cite{rsn} is based on the observation that basic RNNs may be unsuitable for LP, because they do not explicitly handle the path alternation of entities and relations, and when predicting a fact tail, in the current time step they are only passed its relation, and not the head (seen in the previous step). To overcome these issues, they propose Recurrent Skipping Networks (RSNs): in any time step, if the input is a relation, the hidden state is updated re-using the fact head too. The fact score is computed performing the dot product between the output vector and the target embedding.
In training, the model learns relation paths built from the train facts using biased random walk sampling. It employs a specially optimized loss function resorting to a type-based noise contrastive estimation.
In Table~\ref{tab:loss_table} we denote the RSN operation with $rsn$; the number of layers stacked in a RSN cell as $L$; the number of weight matrices as $k$; the number of neurons in each RSN layer as $n$.

\begin{table}[]
    \includegraphics[width=\textwidth]{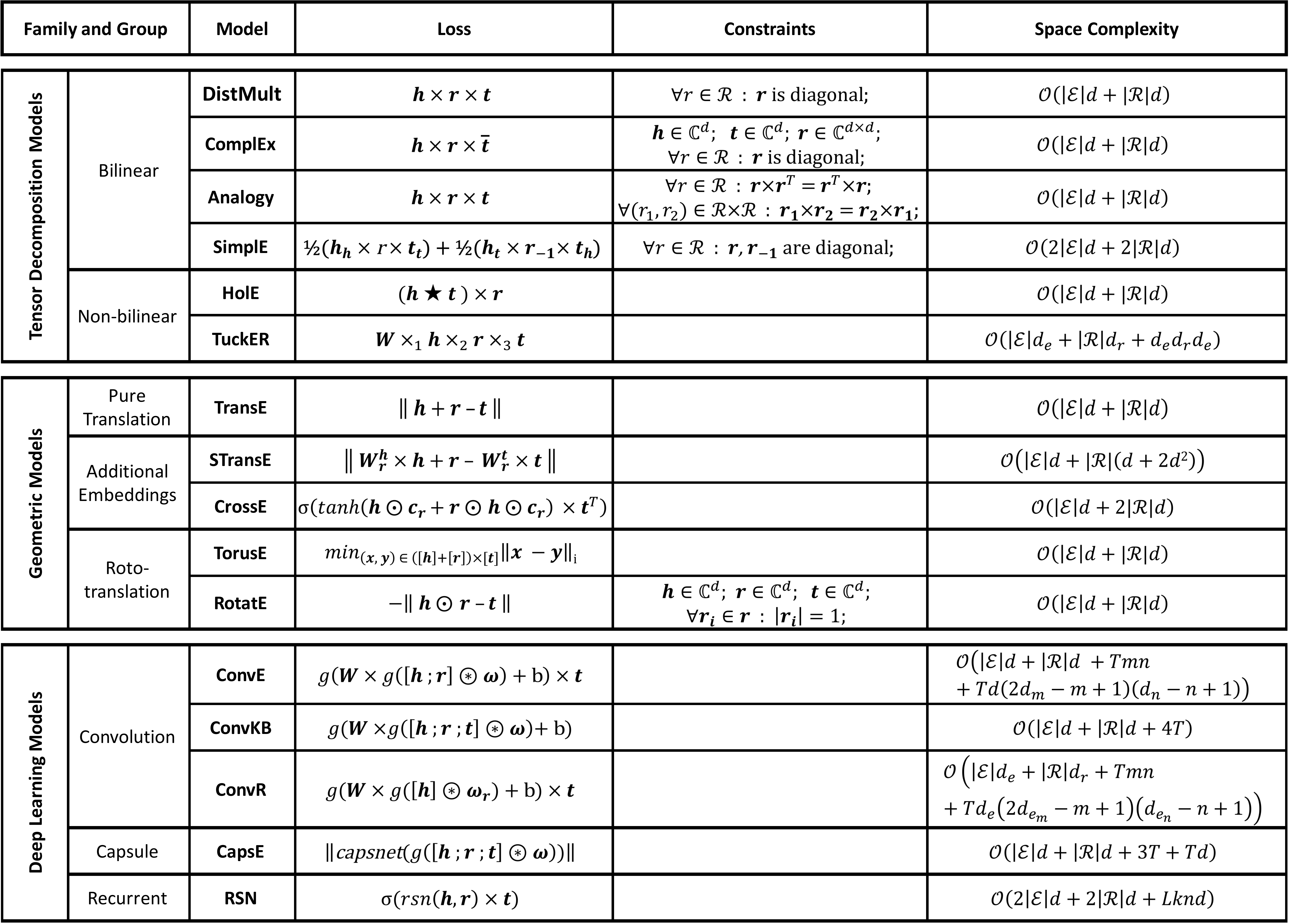}
    \caption{\label{tab:loss_table}
    Loss Function, constraints and space complexity for the models included in our analysis.}
\end{table}

\section{Methodology}
\label{sec:methodology}
In this section we describe the implementations and training protocols of the models discussed before, as well as the datasets and procedures we use to study their efficiency and effectiveness.

\subsection{Datasets}
\label{sec:datasets}
Datasets for benchmarking LP are usually obtained by sampling real-world KGs, and then splitting the obtained facts into a training, a validation and a test set. We conduct our analysis using the 5 best-established datasets in the LP field; we report some of their most important properties in Table~\ref{tab:datasets}.

\mypar{FB15k} is probably the most commonly used benchmark so far. Its creators~\cite{transe} selected all the FreeBase entities with more than 100 mentions and also featured in the Wikilinks database;\footnote{\url{https://code.google.com/archive/p/wiki-links/}} they extracted all facts involving them (thus also including their lower-degree neighbors), except the ones with literals, e.g. dates, proper nouns, etc. They also converted $n$-ary relations represented with reification into cliques of binary edges; this operation has greatly affected the graph structure and semantics, as described in Section~\ref{sec:methodology:reified}. 

\mypar{WN18}, also introduced by the authors of TransE~\cite{transe}, was extracted from WordNet\footnote{\url{https://wordnet.princeton.edu/}}, a linguistig KG ontology meant to provide a dictionary/thesaurus to support NLP and automatic text analysis. In WordNet entities correspond to \emph{synsets} (word senses) and relations represent their lexical connections (e.g. ``hypernym''). In order to build WN18, the authors used WordNet as a starting point, and then iteratively filtered out entities and relationships with too few mentions. 

\mypar{FB15k-237} is a subset of FB15k built by~Toutanova~and~Chen ~\cite{toutanova2015observed}, inspired by the observation that FB15k suffers from \emph{test leakage}, consisting in test data being seen by models at training time. In FB15k this issue is due to the presence of relations that are near-identical or the inverse of one another. In order to assess the severity of this problem, Toutanova and Chen have shown that a simple model based on observable features can easily reach state-of-the-art performance on FB15k. FB15k-237 was built to be a more challenging dataset: the authors first selected facts from FB15k involving the 401 largest relations and removed all equivalent or inverse relations. In order to filter away all trivial triples, they also ensured that none of the entities connected in the training set are also directly linked in the validation and test sets.

\mypar{WN18RR} is a subset of WN18 built by~Dettmers~\emph{et~al}.~\cite{dettmers2018convolutional}, also after observing test leakage in WN18. They demonstrate the severity of said leakage by showing that a simple rule-based model based on inverse relation detection, dubbed Inverse Model, achieves state-of-the-art results in both WN18 and FB15k. To resolve that, they build the far more challenging WN18RR dataset by applying a pipeline similar to the one employed for FB15k-237~\cite{toutanova2015observed}. It has been recently acknowledged by the authors~\cite{impl_conve} that the test set includes 212 entities that do not appear in the training set, making it impossible to reasonably predict about 6.7\% test facts.

\mypar{YAGO3-10}, sampled from the YAGO3 KG~\cite{yago3}, was also proposed by~Dettmers~\emph{et~al}.~\cite{dettmers2018convolutional}. It was obtained selecting entities with at least 10 different relations and gathering all facts involving them, thus also including their neighbors. Moreover, unlike FB15k and FB15k-237, YAGO3-10 also keeps the facts about textual attributes found in the KG. As a consequence, as stated by the authors, the majority of its triples deals with descriptive properties of people, such as citizenship or gender. That the poor performances of the Inverse Model~\cite{dettmers2018convolutional} in YAGO3-10 suggest that this benchmark should not suffer from the same test leakage issues as FB15k and WN18.

\begin{table}[]
    \includegraphics[width=0.6\textwidth]{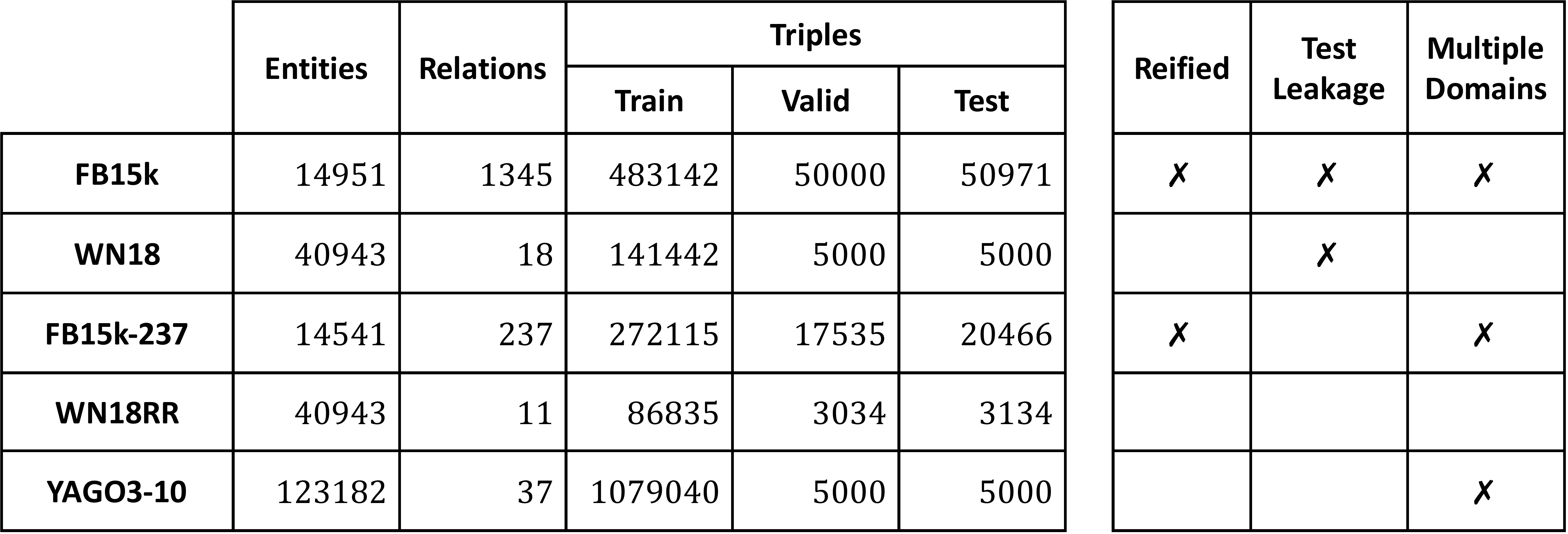}
\caption{The 5 LP datasets included in our comparative analysis, and their general properties.\label{tab:datasets}}
\end{table}

\subsection{Efficiency Analysis}
For each model, we consider two main formulations for efficiency:
\begin{itemize}
\item \emph{Training Time}: the time required to learn the optimal embeddings for all entities and relations.
\item \emph{Prediction Time}: the time required to generate the full rankings for one test fact, including both head and tail predictions.
\end{itemize}
Training Time and Prediction Time mostly depend $(i)$ on the model architecture (e.g. deep neural networks may require longer computations due to their inherently longer pipeline of operations); $(ii)$ on the model hyperparameters, such as embedding size and number of negative samples for each positive one; $(iii)$ on the dataset size, namely the number of entities and relations to learn and, for the Training Time, the number of training triples to process.
Training Time and Prediction Time mostly depend $(i)$ on the model architecture (e.g. deep neural networks may require longer computations due to their shared parameters); $(ii)$ on the model hyperparameters, such as embedding size and number of negative samples for each positive one; $(iii)$ on the dataset size, namely the number of entities and relations to learn and, for the Training Time, the number of training triples to process.

\subsection{Effectiveness Analysis}
We analyze the effectiveness of LP models based on the structure of the training graph. Therefore, we define measurable structural features and we treat each of them as a separate research direction, investigating how it correlates to the predictive performance of each model in each dataset.

We take into account 4 different structural features for each test fact:
\begin{itemize}
    \item \emph{Number of Peers}, namely the valid alternatives for the source and target entities;
    \item \emph{Relational Path Support}, taking into account paths connecting the head and tail of the test fact;
    \item \emph{Relation Properties} that affect both the semantics and the graph structure;
    \item \emph{Degree of the original reified relation}, for datasets generated from KGs using reification.
\end{itemize}

We address these features in Sections~\ref{sec:methodology:peers}, \ref{sec:methodology:path},~\ref{sec:methodology:relprop},~\ref{sec:methodology:reified} respectively.

\subsubsection{Number of Peers}
\label{sec:methodology:peers}
\begin{itemize}
    \item\emph{head peers}: the set of entities $\{h'~\in~\mathcal{E}~\lvert~\langle~h',~r,~t~\rangle~\in~\mathcal{G}_{train}\}$; 
    \item\emph{tail peers}: the set of entities $\{t'~\in~\mathcal{E}~\lvert~\langle~h,~r,~t'~\rangle~\in~\mathcal{G}_{train}\}$.
\end{itemize}
In other words, the head peers are all the alternatives for $h$ seen during training, conditioned to having relation $r$ and tail $t$. Analogously, tail peers are the alternatives for $t$ when the head is $h$ and the relation is $r$. Consistently to the notation introduced in Section~\ref{sec:problem_overview}, we identify the peers for the source and the target entity of a prediction as \textit{source peers} and \textit{target peers} respectively.

\begin{figure}
    \centering{
        \includegraphics[width=0.65\textwidth]{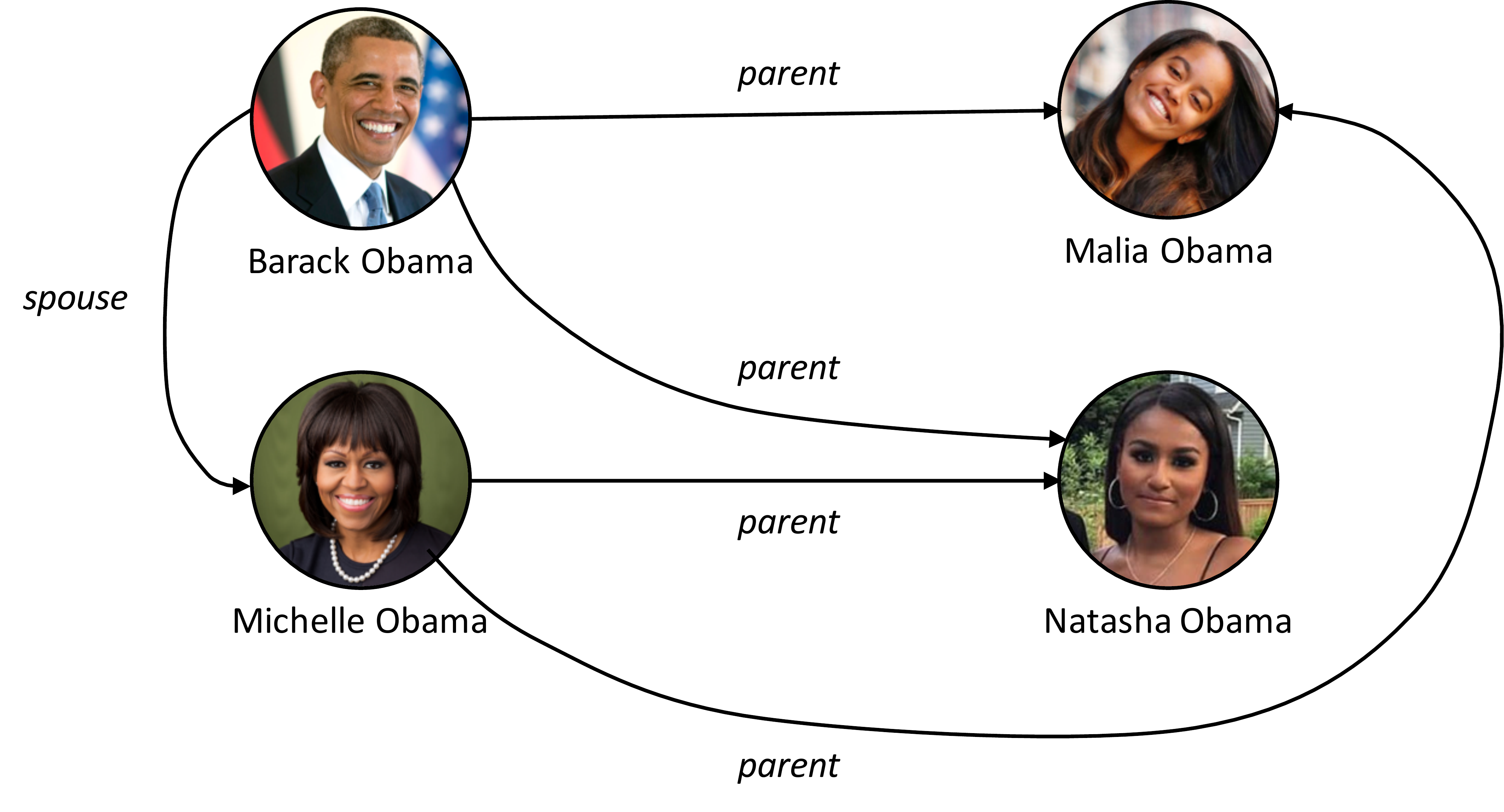}
    }
    \caption{Example of head peers and tail peers in a small portion of a KG.}
    \label{fig:peers}
\end{figure}

We illustrate an example in Figure~\ref{fig:peers}: considering the fact \textlangle{}\textit{Barack~Obama},~\textit{parent},~\textit{Malia~Obama}\textrangle{}, the entity \textit{Michelle~Obama} would be a peer for \textit{Barack~Obama}, because entity \textit{Michelle~Obama} is \textit{parent} to \textit{Malia~Obama} too. Analogously, entity \textit{Natasha~Obama} is a peer for \textit{Malia~Obama}. In head prediction, when \textit{Malia~Obama} is the source entity and \textit{Barack~Obama} is the target entity, \textit{Michelle~Obama} is a target peer and \textit{Natasha~Obama} is a source peer. In tail prediction peers are just reversed: since now \textit{Malia~Obama} is target entity and \textit{Barack~Obama} is source entity, \textit{Michelle~Obama} is a source peer whereas \textit{Natasha~Obama} is a target peer. 

Our intuition is that the numbers of source and target peers may affect predictions with subtle, possibly unanticipated, effects.

On the one hand, the number of source peers can be seen as the number of training samples from which models can directly learn how to predict the current target entity given the current relation. For instance, when performing tail prediction on fact \textlangle{}\textit{Barack~Obama},~\textit{nationality},~\textit{USA}\textrangle{}, the source peers are all the other entities with \textit{nationality} \textit{USA} that the model gets to see in training: they are the examples from which our models can learn what can make a person have American citizenship. 

On the other hand, the number of target peers can be seen as the number of answers correctly satisfying this prediction seen by the model during training. For instance, given the same fact as before \textlangle{}\textit{Barack~Obama},~\textit{nationality},~\textit{USA}\textrangle{}, but performing head prediction this time, the other USA citizens seen in training are now target peers. Since all of them constitute valid alternatives for the target answers, too many target peers may intuitively lead models to confusion and performance degradation.

Our experimental results on source and target peers, reported in Section~\ref{sec:peers_experiments}, confirm our hypothesis.

\subsubsection{Relational Path Support}
\label{sec:methodology:path}

In any KG a \emph{path} is a sequence of facts in which the tail of each fact corresponds to the head of the next one. The \emph{length} of the path is the number of consecutive facts it contains. In what follows, we call the sequence of relation names (ignoring entities) in a path a \emph{relational path}.

Relational paths allow one to identify patterns corresponding to specific relations. For instance, knowing the facts \textlangle{}\textit{Barack~Obama},~\textit{place~of~birth},~\textit{Honolulu}\textrangle{} and \textlangle{}\textit{Honolulu},~\textit{located~in},~\textit{USA}\textrangle{}, it should be possible to predict that \textlangle{}\textit{Barack~Obama},~\textit{nationality},~\textit{USA}\textrangle{}. Paths have been leveraged for a long time by LP techniques based on observable features, such as the Path Ranking Algorithm~\cite{pra2010},\cite{pra2011}. The same cannot be said about models based on embeddings, in which the majority of them learn individual facts separately. Just a few models directly rely on paths, e.g. PTransE~\cite{ptranse} or, in our analysis, RSN~\cite{rsn}; some models do not employ paths directly in training but use them for additional tasks, as the explanation approach proposed by CrossE~\cite{crosse}.

Our intuition is that even models that train on individual facts, as they progressively scan and learn the entire training set, acquire indirect knowledge of its paths as well. As a consequence, in a roundabout way, they may be able to leverage to some extent the patterns observable in paths in order to make better predictions.

Therefore we investigate how the support provided by paths in training can make test predictions easier for embedding-based models. We define a novel measure of \emph{Relational Path Support} (RPS) that estimates for any fact how the paths connecting the head to the tail facilitate their prediction. 
In greater detail, the RPS value for a fact \textlangle{}\textit{h},~\textit{r},~\textit{t}\textrangle{} measures how the relation paths connecting \textit{h} to \textit{t} match those most usually co-occurring with \textit{r}. In models that heavily rely on relation patterns, a high RPS value should correspond to good predictions, whereas a low one should correspond to bad ones.

Our RPS metric is a variant of the TF-IDF statistical measure~\cite{stanford_information_retrieval} commonly used in Information Retrieval. The TF-IDF value of any word $w$ in a document $D$ of a collection $C$ measures both how relevant and how specific $w$ is to $D$, based respectively on the frequency of $w$ in $D$ and on the number of other documents in $C$ including $w$. 
Any document and any keyword-based query can be modeled as a vector with the TF-IDF values of all words in the vocabulary. Given any query $Q$, a TF-IDF-based search engine will retrieve the documents with vectors most similar to the vector of $Q$.

In our scenario we treat each relation path $p$ as a word and each relation $r$ as a document. When a relation path $p$ co-occurs with a relation $r$ (that is, it connects the head and tail of a fact featuring $r$) we interpret this as the word $p$ belonging to the document $r$. 
We treat each test fact $q$ as a query whose keywords are the relation paths connecting its head to the tail. In greater detail, this is the procedure we apply to compute our RPS measure:
\begin{enumerate}
    \item For each training fact $\langle h,r,t\rangle$ we extract from $\mathcal{G}_{train}$ the set of relational paths $p$ leading from $h$ to $t$.
    Whenever in a path a step does not have the correct orientation, we reverse it and mark its relation with the prefix "INV". 
    Our vocabulary $V$ is the set of resulting relational paths. Due to computational constraints, we limit ourselves to relational paths with length equal or lesser than 3.

    \item We aggregate the extracted sets by the relation of the training fact.
    We obtain, for each relation $r$: 
    \begin{itemize}
        \item the number $n_r$ of training facts featuring $r$;
        \item for each relational path $p~\in~V$, the number $n_{r_p}$ of times that $r$ is supported by $p$. Of course, $\forall(r, p)~n_r\geq n_{r_p}$.
    \end{itemize}
    
    \item We compute \emph{Document Frequencies}~(DFs): $\forall\textit{p}\in\textit{V}\quad DF[p]=|\{r~\in~\mathcal{R} : n_{r_p}>0\}|$.
    
    \item We compute \emph{Term Frequencies}~(TFs): 
    $\forall\textit{r}\in\mathcal{R},~\forall\textit{p}\in\textit{V},\quad~TF[r][p]=
    \frac{n_{r_p}}{\sum_{x \in V}{n_{r_x}}}$.

    \item We compute \emph{Inverse Document Frequencies}~(IDFs): 
    $\forall\textit{p}\in\textit{V}\quad IDF[p]=log(\frac{|\mathcal{R}|}{DF[p]})$.
    
    \item For each relation we compute the \textit{TF-IDF} vector:
    $\forall\textit{r}\in\mathcal{R}\quad TFIDF_r= [\forall\textit{p}\in\textit{V}\quad TF[r][p]*IDF[p]~]$.
    
    \item For each test fact $q$ we extract the set of relational paths connecting its head to its tail analogously to point $(1)$.
    
    \item For each $q$ we apply the same formulas seen in points $(3)$ - $(6)$ to compute \emph{DF}, \emph{TF} and \emph{IDF} and the whole \emph{TF-IDF} vector; in all computations we treat each $q$ as if it was an additional document.
    
    \item For each $q$ we compute $RPS$ as the cosine-similarity between its TF-IDF vector and the TD-IDF vector of its relation $r_q$: $RPS(q) = cossim(TF-IDF_q, TF-IDF_{r_q})$.
\end{enumerate}

\begin{figure}
    \includegraphics[width=\textwidth]{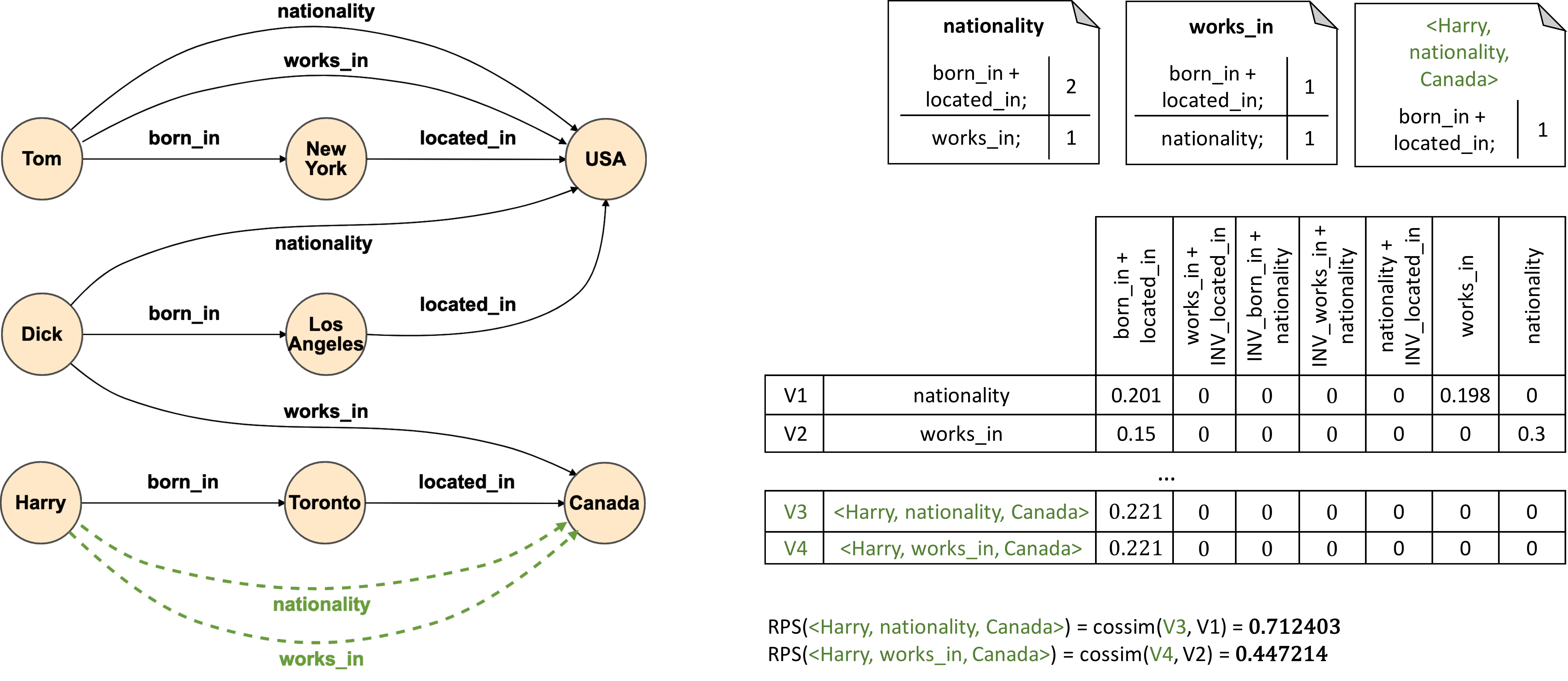}
    \caption{\label{fig:paths}Example for Relational Path Support}
\end{figure}

The RPS of a test fact estimates how similar it is to training facts with the same relation in terms of co-occurring relation paths. This corresponds to measure how much the relation paths suggest that, given the source and relation in the test fact, the target is indeed the right answer for prediction.

\begin{example}

Figure~\ref{fig:paths} shows a graph where black solid edges represent training facts and green dashed edges represent test facts. The collection of documents is $C = \{nationality,$ $works\_in,$ $born\_in,$ $located\_in\}$, and test facts \textlangle{}~\textit{Harry},~\textit{nationality},~ \textit{Canada}~\textrangle{} and \textlangle{}~\textit{Harry},~\textit{works\_in},~ \textit{Canada}~\textrangle{} correspond to two queries. We compute words and frequencies for each document and query. 
Note that the two test facts in our example connect the same head to the same tail, so the corresponding queries have the same keywords (the relational path \textit{born\_in + located\_in}).

We obtain TF-IDF values for each word in each document as described above. For instance, for document $nationality$ and word $born\_in~+~located\_in$:
\begin{itemize}
    \item $TF(born\_in~+~located\_in,~nationality)~=~\frac{co-occurrences~of~born\_in~+~located\_in~with~nationality}{co-occurrences~of~all~relational~ paths~with~nationality}~=\frac{2}{2+1}~\simeq~0.67$
    \item $IDF(born\_in~+~located\_in)~=~\log_{10}(\frac{all~documents}{documents~containing~born\_in~+~located\_in})~=~\log_{10}(\frac{4}{2})~\simeq~0.3$
    \item $TFIDF(born\_in+located\_in,~nationality)~=~TF~\times~IDF~=~0.67*0.3=0.201$
\end{itemize}
Other values can be computed analogously; for instance, $TFIDF(~born\_in~+~located\_in,~works\_in~)=0.15$;\newline$TFIDF(~works\_in,~nationality~)=0.198$.

The TF-IDF value for each query can be computed analogously, except that the query must be included among the documents. The two queries our example share the same keywords, so they will result in identical vectors.
\begin{itemize}
    \item $TF(born\_in+located\_in, test fact)~=~\frac{co-occurrences~of~born\_in~+~located\_in~with~test~fact}{all~relational~paths~co-occurring~with~test~fact}~=\frac{1}{1}~=~1.0$
    \item $IDF(born\_in~+~located)~=~log_{10}(\frac{all~documents}{documents~containing~born\_in~+~located\_in}))~=~log_{10}(\frac{4+1}{2+1})~\simeq~0.221$
    \item $TFIDF(born\_in+located\_in, test fact)~=~TF~\times~IDF~=~1*0.221=0.221$
\end{itemize}

The RPS for \textlangle{}~\textit{Harry},~\textit{nationality},~ \textit{Canada}~\textrangle{} is the cosine-similarity between its vector the vector of \textit{nationality}, and it measures 0.712403; analogously, the RPS for \textlangle{}~\textit{Harry},~\textit{works\_in},~\textit{Canada}~\textrangle{} is the cosine-similarity with the vector of \textit{nationality}, and it measures 0.447214. 
As expected, the former RPS value is higher than the latter: the relational paths connecting \textit{Harry} with \textit{Canada} are more similar to the those usually observed with \textit{nationality} than those usually observed with \textit{works\_in}. In other words, in our small example the relation path \textit{born\_in + located\_in} co-occurs with \textit{nationality} more than with \textit{works\_in}.
\end{example}

While the number of peers only depends on the local neighborhood of the source and target entity, RPS relies on paths that typically have length greater than one. In other words, the number of peers can be seen as a form of information very close to the test fact, whereas RPS is more prone to take into account longer-range dependencies.

Our experimental results on the analysis of relational path support are reported in Section~\ref{sec:rel_path_support_experiments}.

\subsubsection{Relation Properties}
\label{sec:methodology:relprop}
Depending on their semantics, relations can be characterized by several properties heavily affecting the ways in which they appear in the facts. Such properties have been well known in the LP literature for a long time, because they may lead a relation to form very specific structures and patterns in the graph; this, depending on the model, can make their facts easier or harder to learn and predict.

As a matter of fact, depending on their scoring function, some models may be even incapable of learning certain types of relations correctly. For instance, TransE~\cite{transe} and some of its successors are inherently unable to learn symmetric and transitive relations due to the nature of translation itself.
Analogously, DistMult~\cite{distmult} can not handle anti-symmetric relations, because given any fact $\langle h, r, t \rangle$, it assigns the same score to $\langle t, r, h \rangle$ too.

This has led some works to formally introduce the concept of \textbf{full expressiveness}~\cite{simple}: a model is fully expressive if, given any valid graph, there exists at least one combination of embedding values for the model that correctly separates all correct triples from incorrect ones.
A fully expressive model has the theoretical potential to learn correctly any valid graph, without being hindered by intrinsic limitations.
Examples of models that have been demonstrated to be fully expressive are SimplE~\cite{simple}, TuckER~\cite{tucker}, ComplEx~\cite{complex} and HolE~\cite{complex_hole_comparison}. 
 
Being capable of learning certain relations, however, does not necessarily imply reaching good performance on them. Even for fully expressive models, certain properties may be inherently harder to handle than others. For instance, Meilicke~\emph{et~al.}~\cite{rulen2018} have analyzed how their implementations of HolE~\cite{hole}, RESCAL~\cite{rescal} and TransE~\cite{transe} perform on symmetric relations in various datasets; they report surprisingly bad results for HolE on symmetric relations in FB15K, despite HolE being fully expressive).

At this regard, we lead a systematical analysis: we define a comprehensive set of relation properties and verify how they affect performance for all our models.

We take into account the following properties: 
\begin{itemize}
    \item\emph{Reflexivity}: in the original definition, a reflexive relation connects each element with itself. This is not suitable for KGs, where different entities may only be involved with some relations, based on their type. As a consequence, in our analysis we use the following definition: 
    $r \in \mathcal{R}$ is reflexive if 
    $\forall \langle h, r, t \rangle \in \mathcal{G}_{train}$, 
    $\langle h, r, h \rangle \in \mathcal{G}_{train}$ too.

    \item\emph{Irreflexivity}: $r \in \mathcal{R}$ is irreflexive if 
    $\forall e \in \mathcal{E}\quad \langle e, r, r\rangle \notin \mathcal{G}_{train}$.

    \item\emph{Symmetry}: $r \in \mathcal{R}$ is symmetric if 
    $\forall \langle h, r, t \rangle \in \mathcal{G}$, 
    $\langle t, r, h \rangle \in \mathcal{G}$ too.

    \item\emph{Anti-symmetry}: $r \in \mathcal{R}$ is anti-symmetric if 
    $\forall \langle h, r, t \rangle \in \mathcal{G}$, 
    $\langle t, r, h \rangle \notin \mathcal{G}$.

    \item\emph{Transitivity}: $r \in \mathcal{R}$ is transitive if 
    $\forall$ pair of facts $\langle h, r, x \rangle \in \mathcal{G}$ and $\langle x, r, t \rangle \in \mathcal{G}$,
    $\langle h, r, t \rangle \in \mathcal{G}$ as well.
\end{itemize}

We do not consider other properties, such as Equivalence and Order (partial or complete), because we experimentally observed that in all datasets included in our analysis only a negligible number of facts would be included in the resulting buckets.

On each dataset we use the following approach. First, for each relation in the dataset we extract the corresponding training facts and use them to identify its properties. Due to the inherent incompleteness of the training set, we employ a tolerance threshold: a property is verified if the ratio of training facts showing the corresponding behaviour exceeds the threshold. In all our experiments, we set tolerance to $0.5$. 
Then, we group the test facts based on the properties of their relations. If a relation possesses multiple properties, its test facts will belong to multiple groups.
Finally, we compute predictive performance scored by each model on each group of test facts.

We report our results regarding relation properties in Section~\ref{sec:relprops_experiments}.

\subsubsection{Reified Relations}
\label{sec:methodology:reified}
Some KGs support relations with cardinality greater than 2, connecting more than two entities at a time. In relations, cardinality is closely related to semantics, and some relations inherently make more sense when modeled in this way. For example, an actor winning an award for her performance in a movie can be modeled with a unique relation connecting the actor, the award and the movie. KGs that support relations with cardinality greater than 2 often handle them in one of the following ways:
\begin{itemize}
    \item using \textbf{hyper-graphs}: in a hyper-graph, each hyper-edge can link more than two nodes at a time by design. Hyper-graphs can not be directly expressed as a set of triples.
    \item using \textbf{reification}: if a relation needs to connect multiple entities, it is modeled with an intermediate node linked to those entities by binary relations. The relation cardinality thus becomes the degree of the reified node. Reification allows relations with cardinality greater than 2 to be indirectly modeled; the graph is thus still representable as a set of triples.
\end{itemize}

The popular KG FreeBase, that has been used to generate important LP datasets such as FB15k and FB15k-237, employs reified relations, with intermediate nodes of type \textit{Compound Value Type} (CVT). By extension, we refer to such intermediate nodes as CVTs.

\begin{figure}
\includegraphics[width=\textwidth]{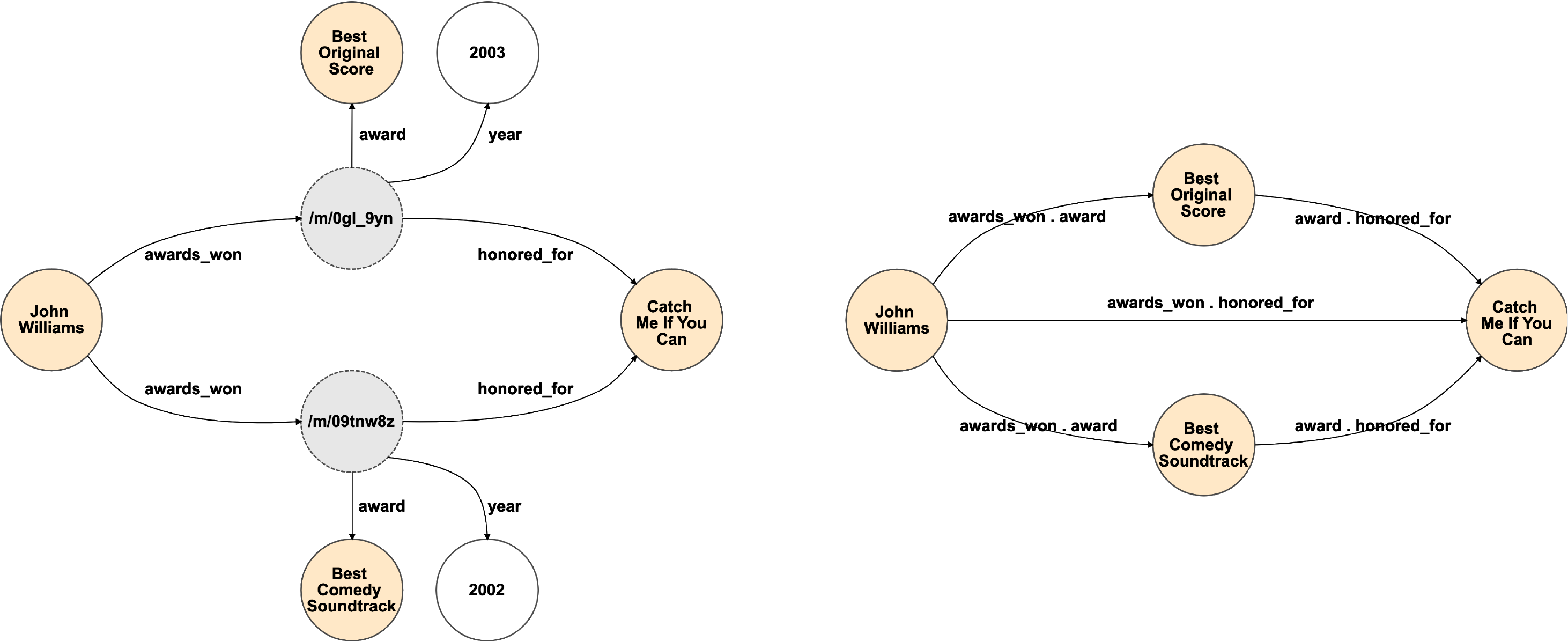}
    \caption{\label{fig:s2c}Example of how the Star2Clique process operates on a small portion of a KG.}
\end{figure}

In the process of extracting FB15k from FreeBase~\cite{transe}, CVTs were removed and converted into cliques in which the entities previously connected to the CVT are now connected to one another; the labels of the new edges are obtained concatenating the corresponding old ones. This also applies to FB15k-237, that was obtained by just sampling FB15k further~\cite{toutanova2015observed}. 
It has been pointed out that this conversion, dubbed ``Star-to-Clique'' (S2C), is irreversible~\cite{wen2016binary}. In our study we have observed further consequences to the S2C policy:
\begin{itemize}
    \item From a structural standpoint, S2C transforms a CVT with degree $n$ into a clique with at most at most $n!$ edges. Therefore, some parts of the graph become locally much denser than before. The generated edges are often redundant, and in the filtering operated to create FB15k-237, many of them are removed. 
    \item From a semantic standpoint, the original meaning of the relations is vastly altered. After exploding CVTs into cliques, deduplication is performed: if the same two entities were linked multiple times by the same types of relation using multiple CVTs -- e.g. an artist winning multiple awards for the same work -- this information is lost, as shown in Figure~\ref{fig:s2c}. In other words, in the new semantics, each fact has happened \textit{at least once}. 
\end{itemize}

We hypothesize that the very dense, redundant and locally-consistent areas generated by S2C may have consequences on predictive behaviour. Therefore, for FB15k and FB15k-237, we have tried to extract for each test fact generated by S2C the degree of the original CVT, in order to correlate it with the predictive performance of our models.

For each test fact generated by S2C we have tried to recover the corresponding CVT from the latest FreeBase dump available.\footnote{https://developers.google.com/freebase}
As already pointed out, S2C is not reversible due to its inherent deduplication; therefore, this process often yields multiple CVTs for the same fact. In this case, we have taken into account the CVT with highest degree. We also report that, quite strangely, for a few test facts built with S2C we could not find any CVTs in the FreeBase dump. For these facts, we have set the original reified relation degree to the minimum possible value, that is 2.

In this way, we were able to map each test fact to the degree of the corresponding CVT with highest degree; we have then proceeded to investigate the correlations between such degrees and the predictive performance of our models. 

Our results on the analysis of reified relations are reported in Section~\ref{sec:reified_experiments}.

\section{Experimental Results}
\label{sec:experiments}
In this section we provide a detailed report for the experiments and comparisons carried out in our work.

\subsection{Experimental set-up} In this section we provide a brief overview of the environment we have used in our work and of the procedures followed to train and evaluate all our LP models. We also provide a description for the baseline model we use in all the experiments of our analysis.

\subsubsection{Environment}
All of our experiments, as well as the training and evaluation of each model, have been performed on a server environment using 88 CPUs Intel Core(TM) i7-3820 at 3.60GH, 516GB RAM and 4 GPUs NVIDIA Tesla P100-SXM2, each with 16GB VRAM. The operating system is Ubuntu 17.10 (Artful Aardvark). 

\subsubsection{Training procedures}
We have trained and evaluated from scratch all the models introduced in Section~\ref{sec:taxonomy}. In order to make our results directly reproducible, we have employed, whenever possible, publicly available implementations. As a matter of fact we only include one model for which the implementation is not available online, that is ConvR~\cite{convr}; we thank the authors for kindly sharing their code with us.

When we have found multiple implementations for the same model, we have always chosen the best best performing one, with the goal of analyzing each model at its best. This has resulted in the following choices:
\begin{itemize}
    \item For TransE~\cite{transe}, DistMult~\cite{distmult} and HolE~\cite{hole} we have used the implementation provided by project Ampligraph~\cite{ampligraph} and available in their repository~\cite{impl_ampligraph};
    \item For ComplEx~\cite{complex} we have used Timoth\'ee Lacroix's version with N3 regularization~\cite{complex_n3}, available in Facebook Research repository~\cite{impl_complex};
    \item For SimplE ~\cite{simple} we have used the fast implementation by Bahare Fatemi~\cite{impl_simple}, as suggested by the creators of the model themselves.
\end{itemize}
For all the other models we use the original implementations shared by the authors themselves in their repositories.

As shown by Kadlec~\emph{et~al}.~\cite{kadlec2017knowledge}, LP models tend to be extremely sensitive to hyperparameter tuning, and the hyperparameters for any model often need to be tuned separately for each dataset. 
The authors of a model usually define the space of acceptable values of each hyperparameter, and then run grid or random search to find the best performing combination. 

In our trainings, we have relied on the hyperparameter settings reported by the authors for all datasets on which they have run experiments. Not all authors have evaluated their models on all of the datasets we include in our analysis, therefore in several cases we have not found official guidance on the hyperparameters to use. In these cases, we have explored ourselves the spaces of hyperparameters defined by the authors in their papers.

Considering the sheer size of such spaces (often containing thousands of combinations), as well as the duration of each training (usually taking several hours), running a grid search or even a random search was generally unfeasible.
We have thus resorted to hand tuning to the best of our possibilities, using what is familiarly called a \textit{panda} approach~\cite{pandacaviarbook} (in contrast to a \textit{caviar} approach where large batches of training processes are launched). We report in Appendix~\ref{appendix:hyperparams} the best hyperparameter combination we have found for each model, in Table~\ref{tab:hyperparams_table}. 

The filtered H@1, H@10, MR and MRR results obtained for each model in each dataset are displayed in Table~\ref{tab:performances_table}. As mentioned in Section~\ref{sec:methodology:metrics}, for models relying on \textit{min} policy in their original implementation we report their results obtained with \textit{average} policy instead, as we observed that \textit{min} policy can lead to results not directly comparable to those of the other models. We investigate this phenomenon in Section~\ref{sec:experiments:tie_policy}
We note that in AnyBURL~\cite{anyburl2019}, for datasets FB15k-237 and YAGO3-10 we used training time 1,000 secs whereas the original papers report slightly better results with a training time of 10000s; this is because, due to our already described necessity for full rankings, when using models trained for 10,000 secs prediction times got prohibitively long. Therefore, under suggestion of the authors themselves, we resorted to using the second best training time 1,000 secs for these datasets. 
We also note that STransE~\cite{stranse}, ConvKB~\cite{convkb} and CapsE~\cite{capse} use transfer learning and require embeddings pre-trained on TransE~\cite{transe}. For FB15k, FB15k-237, WN18 and WN18RR we have found and used TransE~\cite{transe} embeddings trained and shared by the authors themselves across their repositories; for YAGO3-10, on which the authors did not work, we used embeddings trained with the TransE implementation that we used in our analysis~\cite{impl_ampligraph}.

\begin{table}
    \centering{
        \includegraphics[width=\textwidth]{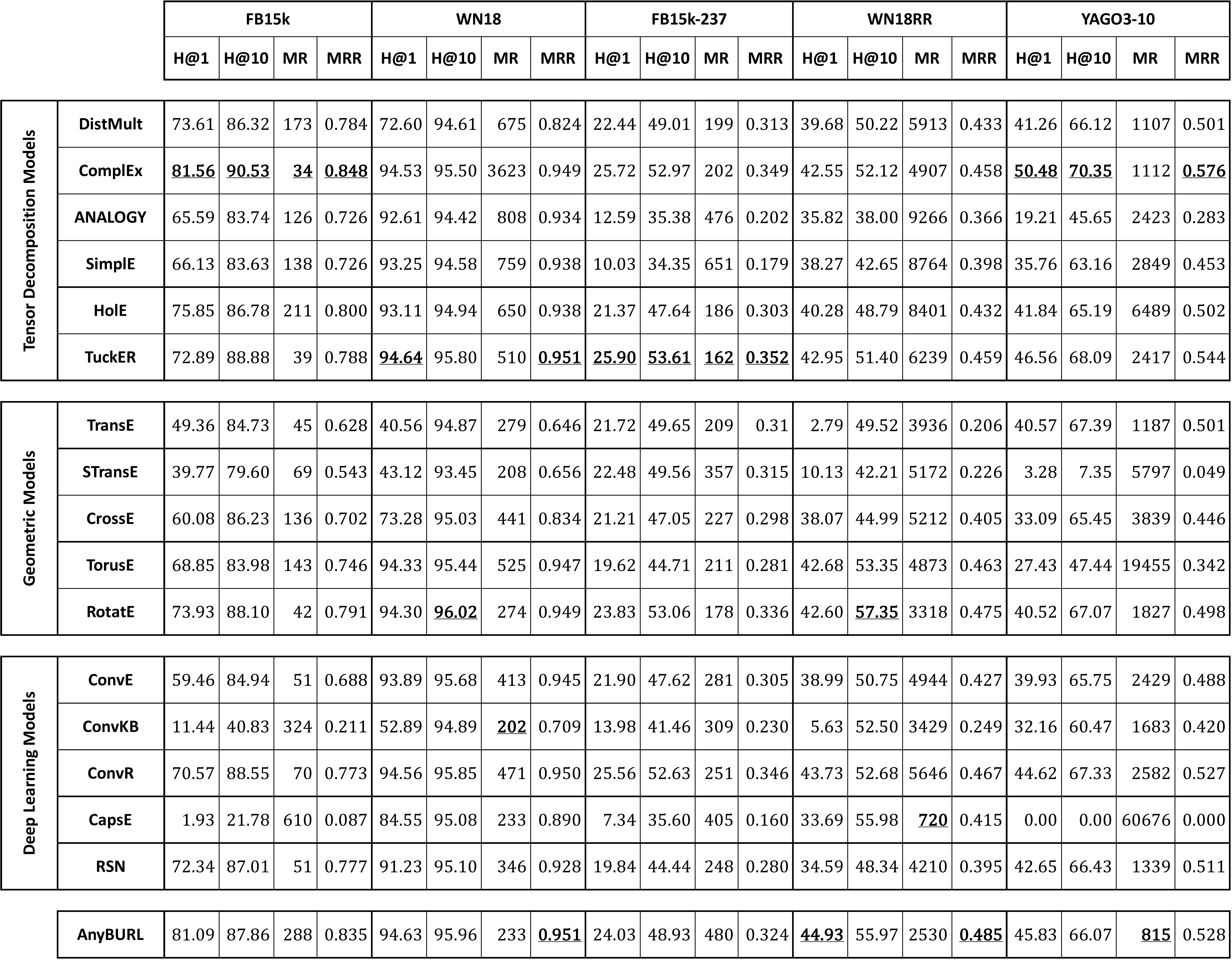}
        \caption{\label{tab:performances_table} Global H@1, H@10, MR and MRR results for all LP models on each dataset. The best results of each metric for each dataset are marked in bold and underlined.}
    }
\end{table}

\subsubsection{Evaluation Metrics}
\label{sec:methodology:metrics}
When it comes to evaluate the predictive performance of models, we focus on the \textbf{filtered scenario}, and, when reporting global results, we use all the four most popular metrics: H@1, H@10, MR and MRR.  

In our finer-grained experiments, we have run all our experiments computing both H@1 and MRR. The use of a H@K measure coupled with a mean-rank-based measure is very popular in LP works. 
At this regard, we focus on H@1 instead of using larger values of $K$ because, as observed by Kadlec~\emph{et~al}.~\cite{kadlec2017knowledge}, low values of $K$ allow the emerging of more more marked differences among different models. 
Similarly, we choose MRR because is a very stable metric, while simple MR tends to be highly sensitive to outliers. 
In this paper we mostly report H@1 results for our experiments, as we have usually observed analogous trends using MRR.

As described in Section~\ref{sec:problem_overview}, we have observed that the implementations of different models may rely on different policies for handling ties. Therefore, we have modified them in order to extract evaluation results with multiple policies for each model. In most cases we have not found significant variations; nonetheless, for a few models, \textit{min} policy yields significantly different results from the other policies. In other words, using different tie policies may make results not directly comparable to one another.

Therefore, unless differently specified, for models that employed \textit{min} policy in their original implementation we report their \textit{average} results instead, as the latter are directly comparable to the results of the other models, whereas the former are not. 
We have led further experiments on this topic and report interesting findings in Section~\ref{sec:experiments:tie_policy}

\subsubsection{Baseline}
As a baseline we use AnyBURL~\cite{anyburl2019}, a popular LP model based on observable features. AnyBURL (acronym for Anytime Bottom-Up Rule Learning) treats each training fact as a compact representation of a very specific rule; it then tries to generalize it, with the goal of covering and satisfying as many training facts as possible. In greater detail, AnyBURL samples paths of increasing length from the training dataset. For each path of length $n$, it computes a rule containing $n-1$ atoms, and stores it if some quality criteria are matched. AnyBURL keeps analyzing paths of same length $n$ until a saturation threshold is exceeded; when this happens, it moves on to paths of length $n+1$.

As a matter of fact, AnyBURL is a very challenging baseline for LP, as it is shown to outperform most latent-features-based models. It is also computationally fast: depending on the training setting, in order to learn its rules it can take from a couple of minutes (\emph{100s} setting) to about 3 hours (\emph{10000s} setting).
When it comes to evaluation, AnyBURL is designed to return the top-k scoring entities for each prediction in the test set. When used in this way, it is strikingly fast as well. In order to use it as a baseline in our experiments, however, we needed to extract \textbf{full ranking} for each prediction, setting $k= \lvert \mathcal{E} \lvert$: this resulted in much longer computation times. 

As a side note, we observe that in AnyBURL predictions, even the full ranking may not contain all entities, as it only includes those supported by at least one rule in the model. This means that in very few facts, the target entity may not be included even in the full ranking. In this very unlikely event, we assume that all the entities that have not been predicted have identical score 0, and we apply the \emph{avg} policy for ties. 

Code and documentation for AnyBURL are publicly available~\cite{impl_anyburl}.

\subsection{Efficiency}
In this section we report our results regarding the efficiency of LP models in terms of time required for training and for prediction.

\begin{figure}
    \centering
    {\includegraphics[width=\textwidth]{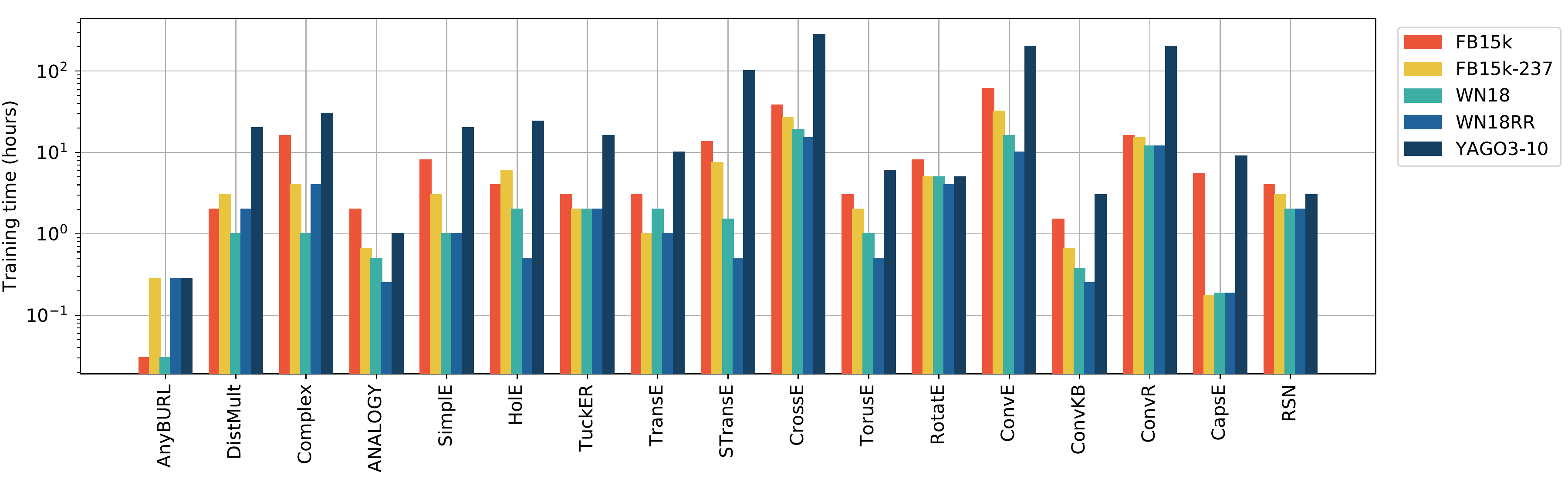}}
\caption{\label{fig:figure_efficiency_training_times} Training times in hours for each LP model on each dataset. Y axis is in logscale.}
\end{figure}

\begin{figure}
    \centering
    {\includegraphics[width=\textwidth]{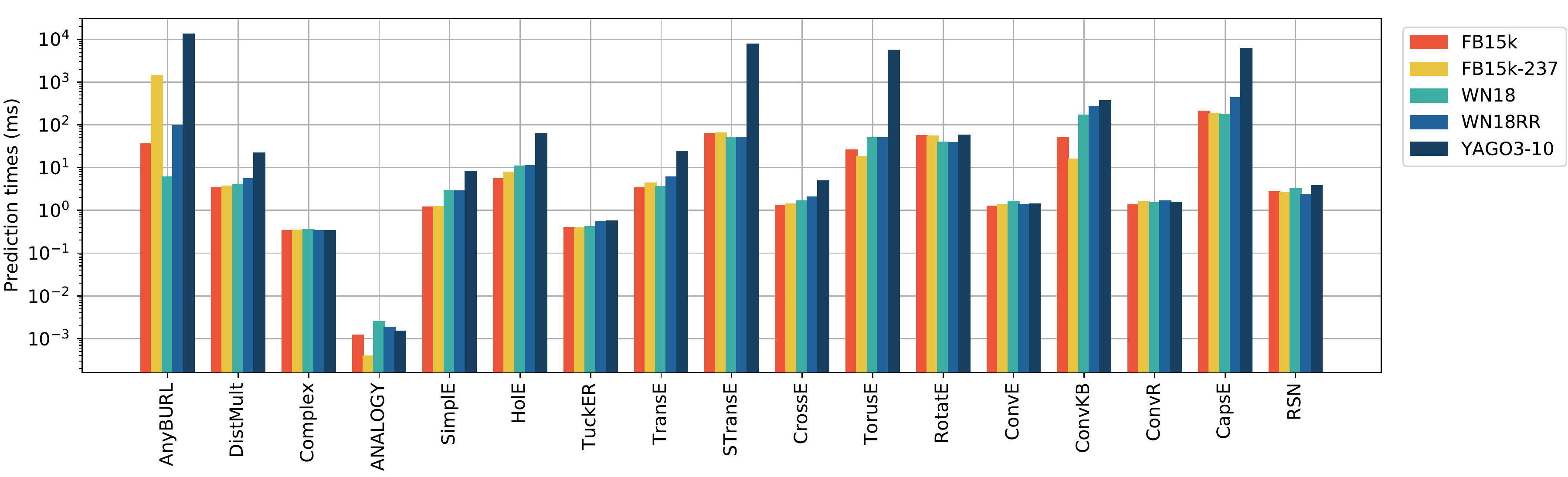}}
\caption{\label{fig:figure_efficiency_scoring_times} Prediction times in milliseconds for each LP model on each dataset. Y axis is in logscale.}
\end{figure}

In Figure~\ref{fig:figure_efficiency_training_times} we illustrate, for each model, the time in hours spent for training on each dataset. We observe that training times range from around 1h to about 200-300 hours. Not surprisingly, the largest dataset YAGO3-10 usually requires significantly longer training times. In comparison to the embedding-based models, the baseline project AnyBURL~\cite{anyburl2019} is strikingly fast. AnyBURL treats the training time as a configuration parameter, and reaches state-of-the-art performance after just 100s for FB15k and WN18, and 1000s for FB15k237, WN18RR and YAGO3-10. As already mentioned, STransE~\cite{stranse}, ConvKB~\cite{convkb} and CapsE~\cite{capse} require embeddings pre-trained on TransE~\cite{transe}; we do not include pre-training times in our measurements. 
In Figure~\ref{fig:figure_efficiency_scoring_times} we illustrate, for each model, the prediction time, defined as the time required to generate the full ranking in both head and tail predictions for one fact. These scores are mainly affected by the embedding dimensions and by the evaluation batch size; at this regard we note that ConvKB~\cite{convkb} and CapsE~\cite{capse} are the only models that require multiple batches for running one prediction, and this may have negatively affected their prediction performance. In our experiments for these models we have used evaluation batch size 2048 (the maximum allowed in our setting).
We observe that ANALOGY behaves particularly well in terms of prediction time; this may possibly depend on this model being implemented in C++. For the baseline AnyBURL~\cite{anyburl2019}, we have obtained prediction times a posteriori by dividing the we whole rules application times by the numbers of facts in the datasets. The obtained prediction times are significantly higher than the ones for embedding-based methods; we stress that this depends on the fact that AnyBURL is not designed to generate full rankings, and that using top-k policy lower $k$ values would result in much faster computations.

\subsection{Effectiveness}
In this section we report our results regarding the effectiveness of LP models in terms of time predictive performance.

\subsubsection{Peer Analysis}
\label{sec:peers_experiments}

Our goal in this experiment is to analyze how the predictive performance varies when taking into account test facts with different numbers of source peers, or tail peers, or both.

\begin{figure}
\centering
    \subfloat[FB15k\label{fig:fb15k_peers}]{\includegraphics[width=\columnwidth]{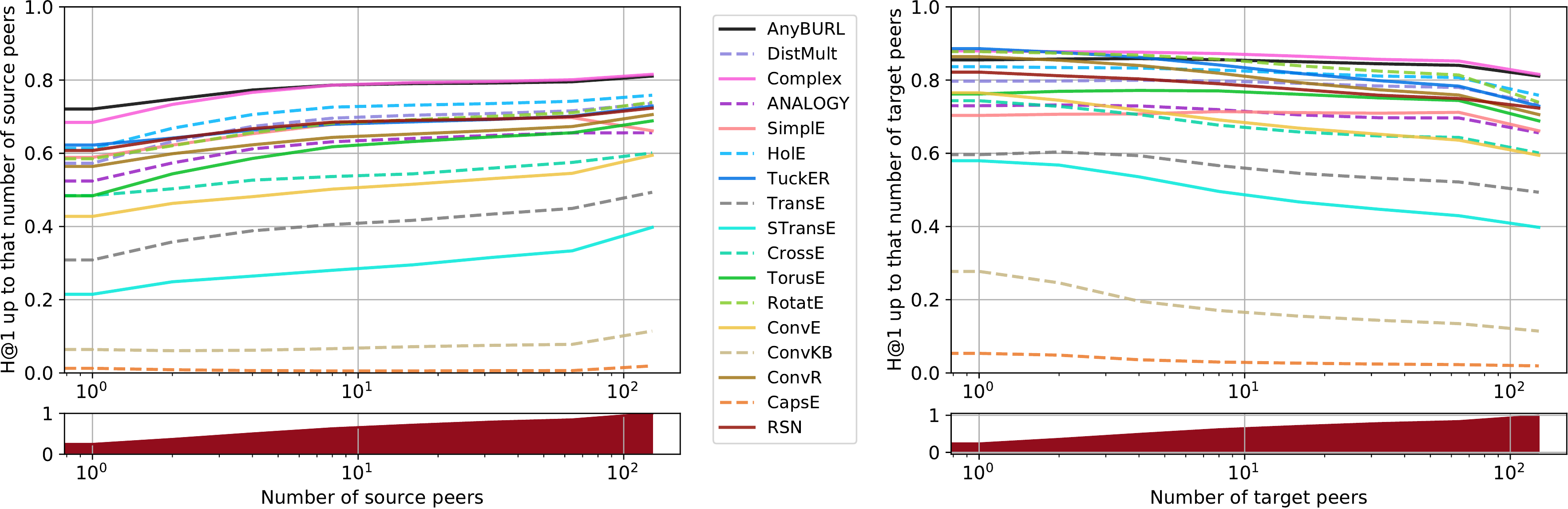}}

    \subfloat[FB15k-237\label{fig:fb15k237_peers}]{\includegraphics[width=\columnwidth]{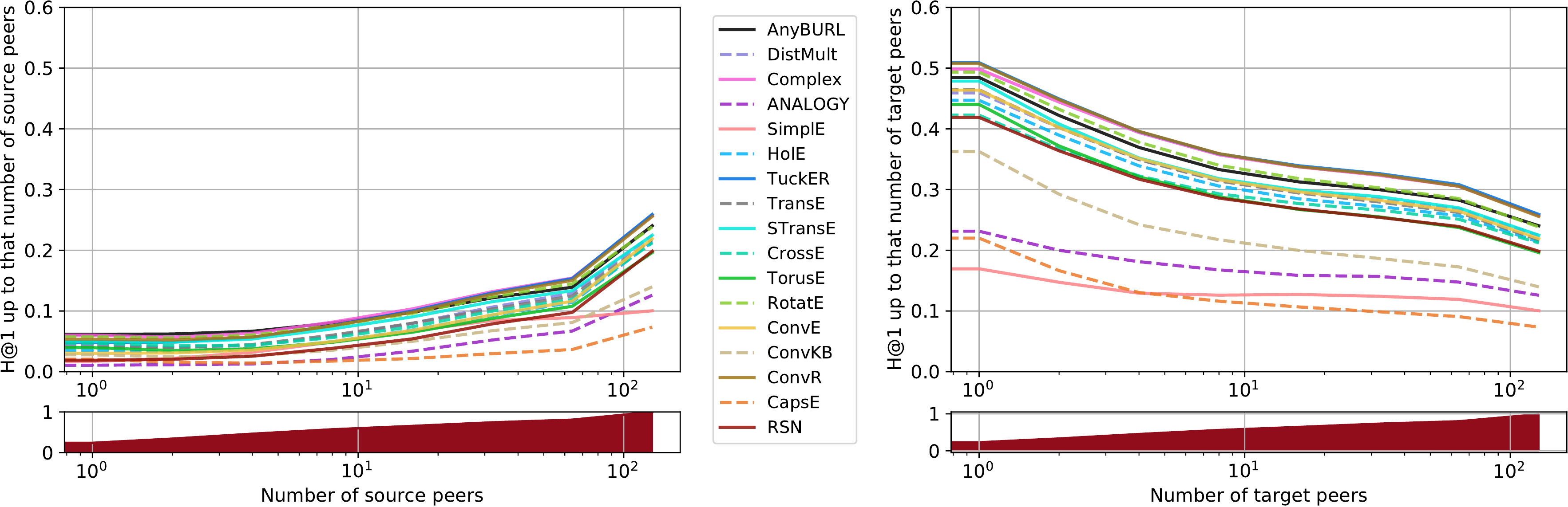}}
    
    \caption{Cumulative H@1 results for each LP model on the \textbf{Freebase} datasets, and corresponding cumulative distribution of test facts, varying the number of source peers (left) and target peers (right). X axis is in logscale.}
\end{figure}

\begin{figure}
\centering
    \subfloat[WN18\label{fig:wn18_peers}]{\includegraphics[width=\columnwidth]{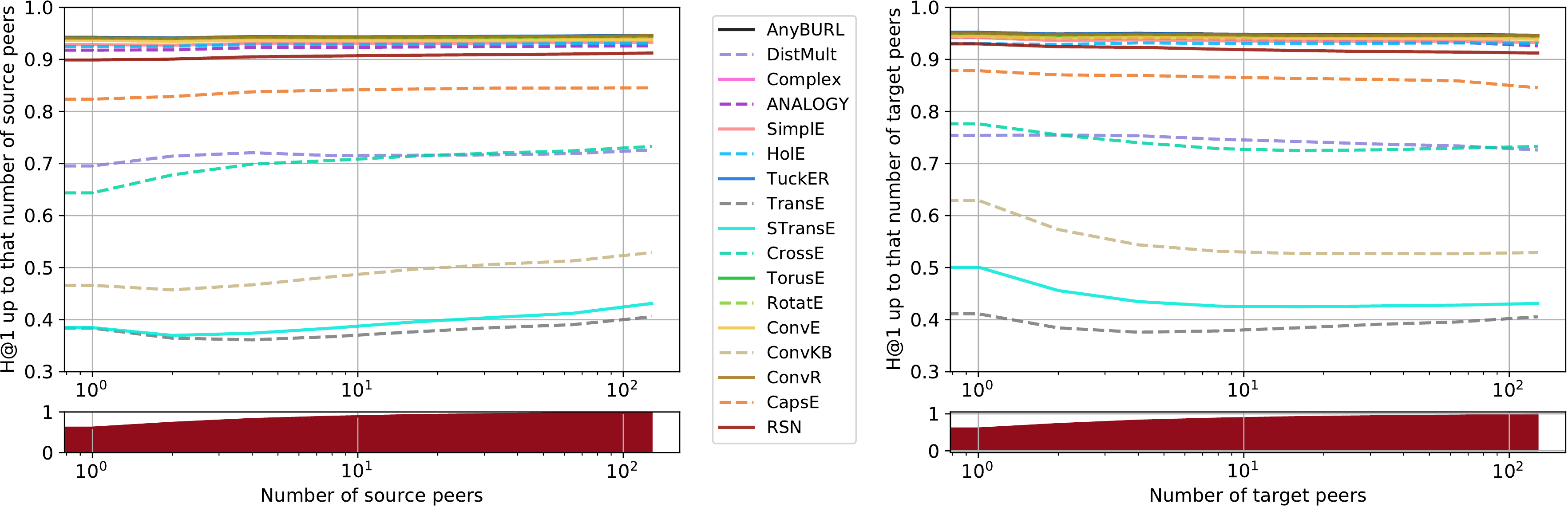}}

    \subfloat[WN18RR\label{fig:wn18rr_peers}]{\includegraphics[width=\columnwidth]{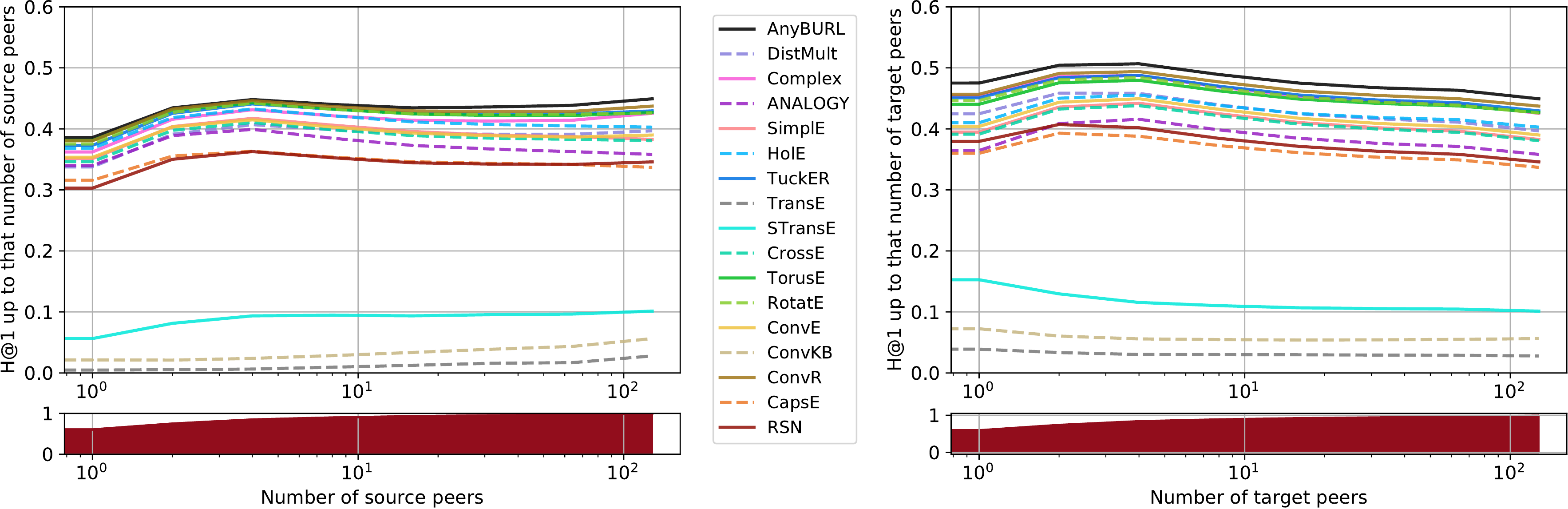}}
    
    \caption{Cumulative H@1 results for each LP model on the \textbf{Wordnet} datasets, and corresponding cumulative distribution of test facts, varying the number of source peers (left) and target peers (right). X axis is in logscale.}
\end{figure}

\begin{figure}
\centering
    \subfloat[YAGO3-10]{\includegraphics[width=\columnwidth]{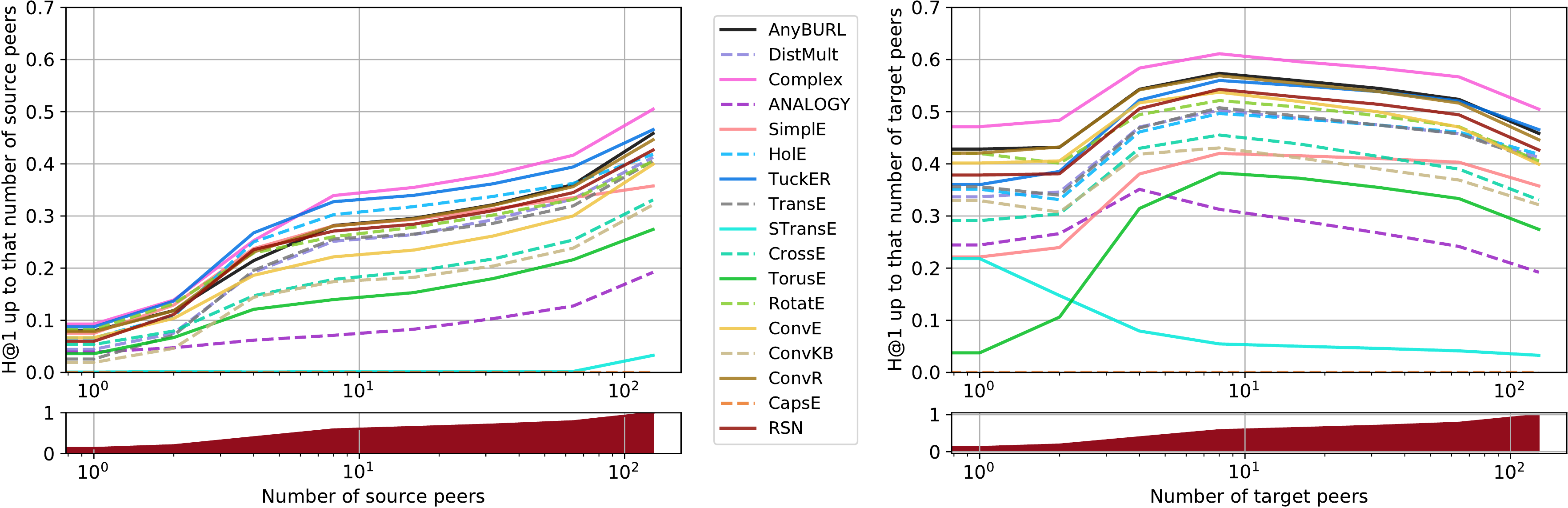}}
    
    \caption{Cumulative H@1 results for each LP model on YAGO3-10, and corresponding cumulative distribution of test facts, varying the number of source peers (left) and target peers (right). X axis is in logscale.\label{fig:yago_peers}}
\end{figure}

We report in Figures \ref{fig:fb15k_peers}, \ref{fig:fb15k237_peers}, \ref{fig:wn18_peers}, \ref{fig:wn18rr_peers}, \ref{fig:yago_peers} our results. These plots show how performances trend when increasing the number of source peers or target peers. We use use H@1 for measuring effectiveness. The plots are incremental, meaning that, for each number of source (target) peers, we report H@1 percentage for all predictions with source (target) peers equal or lesser than that number. In each graph, for each number of source (target) peers we also report the overall percentage of involved test facts: this provides the distribution of facts by peers.

Our observations are intriguing.
First, we point out that almost always, predictions with a greater number of source peers show better H@1 results. A way to explain this phenomenon is to consider that the source peers of a prediction are the examples seen in training in which the target entity displays the same role as in the fact to predict. For instance, when performing tail prediction for \textlangle{}~\textit{Barack~Obama},~\textit{nationality},~\textit{USA}~\textrangle{}, having many source peers means that the model has seen in training numerous examples of people with \textit{nationality} \textit{USA}. Intuitively, such examples provide the models with meaningful information, allowing them to more easily understand when other entities (such as \textit{Barack~Obama)} have  \textit{nationality} \textit{USA} as well.

Second, we observe that very often a greater number of target peers leads to worse H@1 results. For instance, when performing head prediction for \textlangle{}~?~\textit{Michelle~Obama},~\textit{born~in},~\textit{Chicago}~\textrangle{}, target peers are numerous if we have already seen in training many other entities born in Chicago. These entities seem confuse models when they are asked to predict that other entities (such as \textit{Michelle Obama}) are born in \textit{Chicago} as well. 

We underline that this decrease in performance is not caused by the target peers just outscoring the target entity: we are taking into account filtered scenario results, therefore target peers, being valid answers to our predictions, do not contribute to the rank computation.

These correlations between numbers of peers and performance are particularly evident in datasets FB15K and FB15K-237. Albeit at a lesser extent, they are also visible in YAGO3-10, especially regarding the source peers. 
In WN18RR these trends seem much less evident. This is probably due to the very skewed dataset structure: more than 60\% predictions involve less than 1 source peer or target peer. In WN18, where the distribution is very skewed as well, models show pretty balanced behaviours. Most of them reach almost perfect results, above 90\% H@1.

\subsubsection{Relational Path Support}
\label{sec:rel_path_support_experiments}

Our goal in this experiment is to analyze how the predictive effectiveness of LP models varies when taking into account predictions with different values of Relational Path Support (RPS). RPS is computed using the TF-IDF-based metric introduced in Section~\ref{sec:methodology}, using relational paths with length up to 3.

\begin{figure}
    \subfloat[FB15k\label{fig:fb15k_paths}]{\includegraphics[width=\columnwidth]{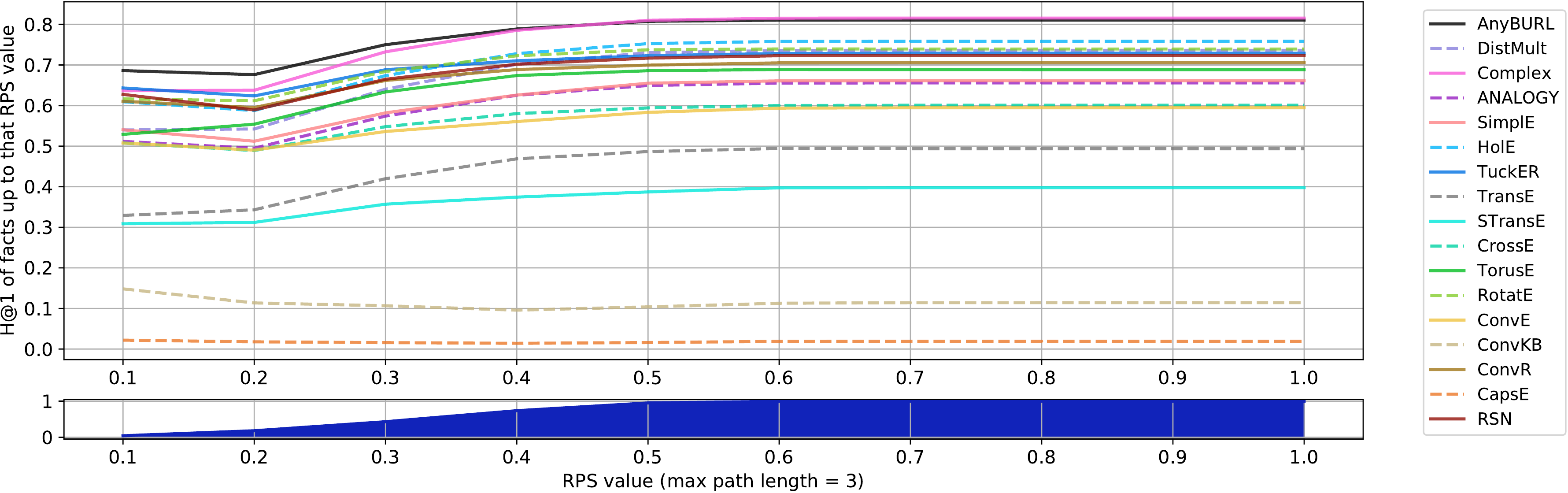}}

    \subfloat[FB15k-237\label{fig:fb15k237_paths}]{\includegraphics[width=\columnwidth]{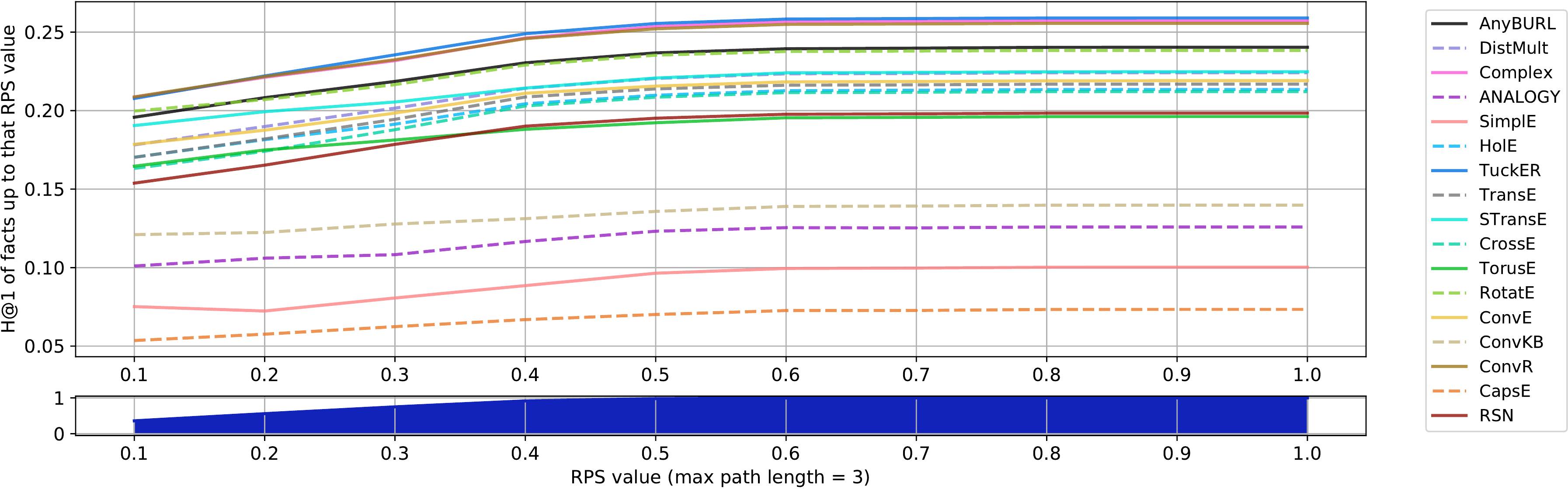}}
    
    \caption{H@1 results for each LP model on the \textbf{Freebase} datasets varying the RPS of the test facts, and corresponding cumulative distribution of test facts.}
\end{figure}

\begin{figure}
    \subfloat[WN18\label{fig:wn18_paths}]{\includegraphics[width=\columnwidth]{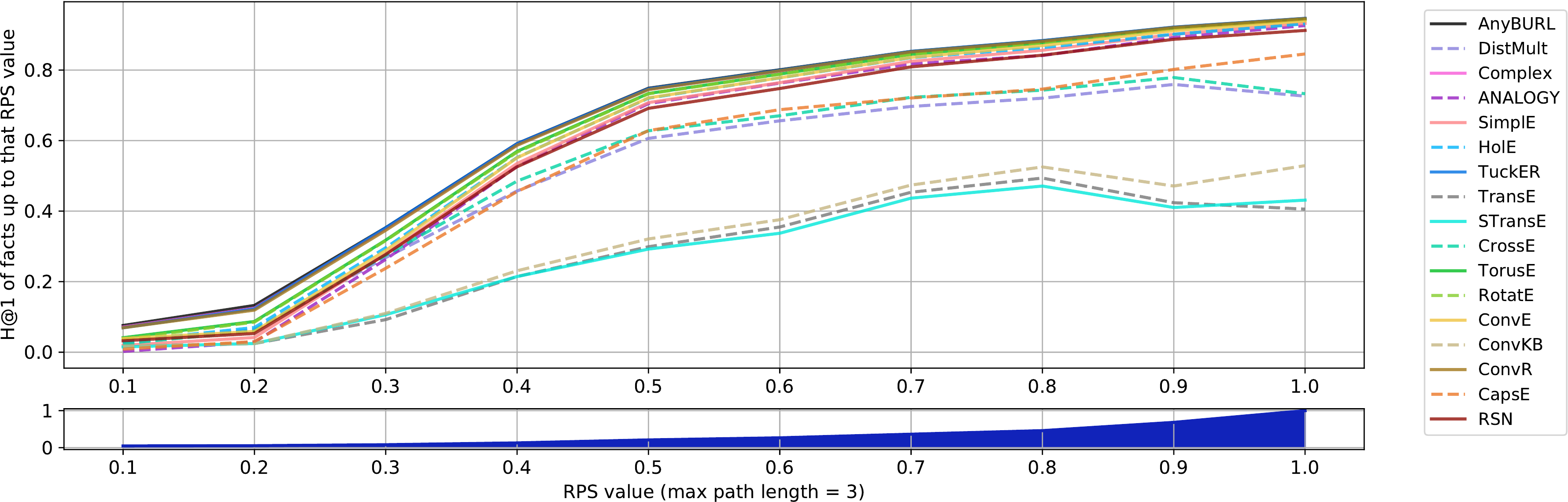}}

    \subfloat[WN18RR\label{fig:wn18rr_paths}]{\includegraphics[width=\columnwidth]{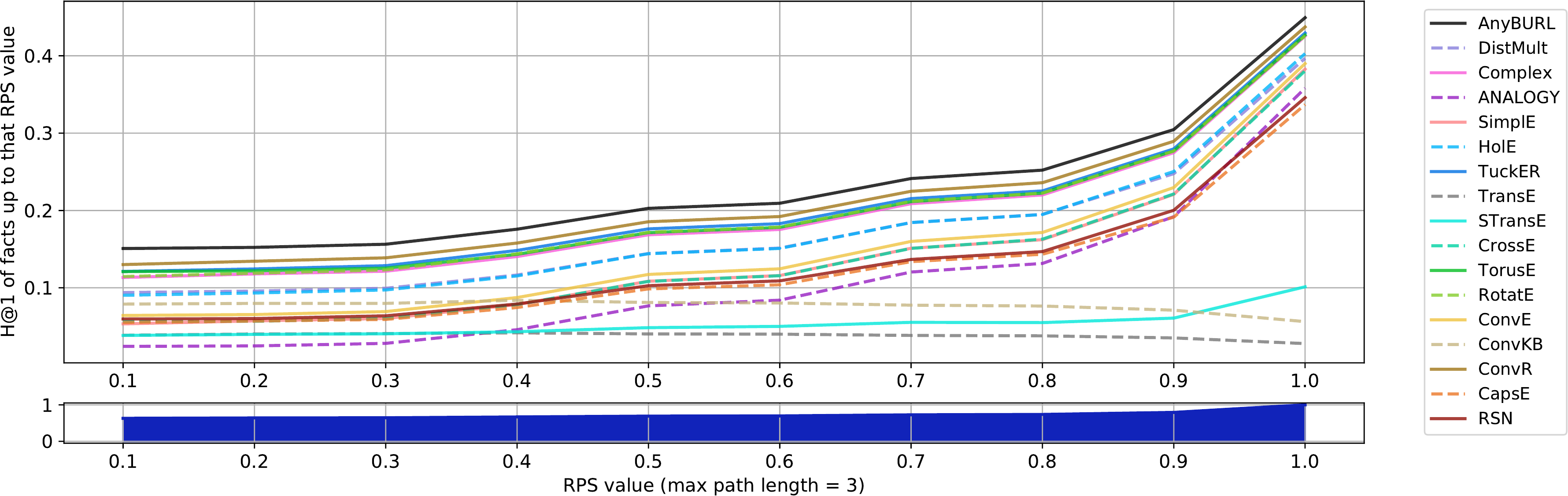}}
    
    \caption{H@1 results for each LP model on the \textbf{Wordnet} datasets varying the RPS of the test facts, and corresponding cumulative distribution of test facts.}
\end{figure}

\begin{figure}
    \subfloat[YAGO3-10]{\includegraphics[width=\columnwidth]{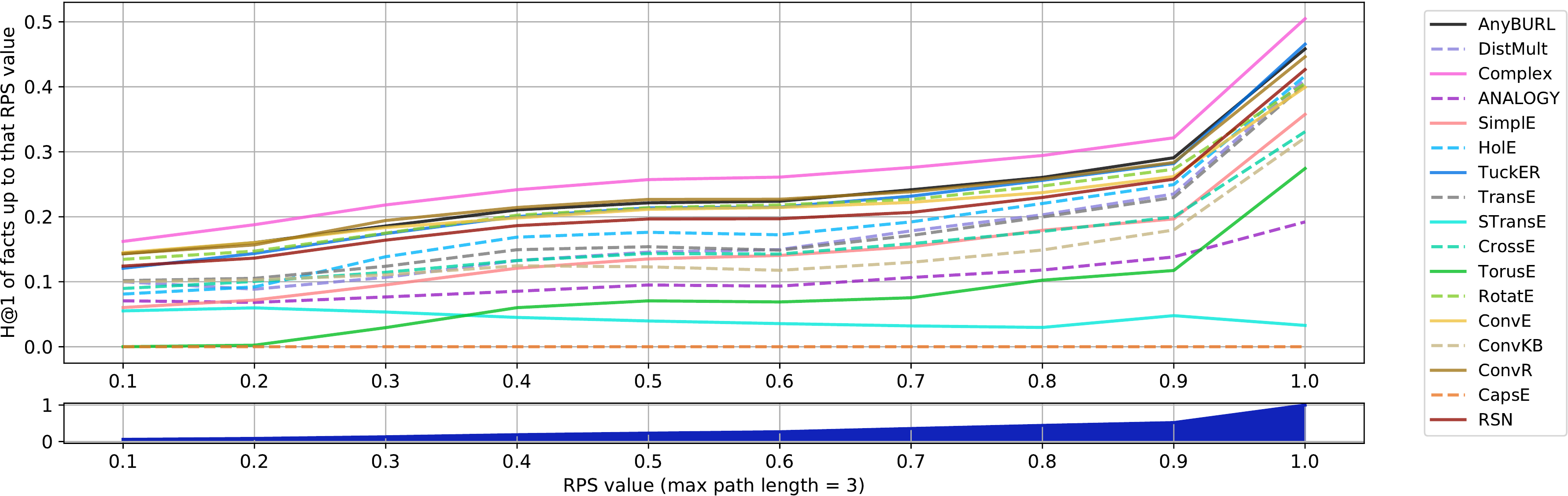}}
    \caption{H@1 results for each LP model on YAGO3-10 varying the RPS of the test facts, and corresponding cumulative distribution.\label{fig:yago_paths}}
\end{figure}

We report in Figures~\ref{fig:fb15k_paths}, \ref{fig:fb15k237_paths}, \ref{fig:wn18_paths}, \ref{fig:wn18rr_paths}, \ref{fig:yago_paths} our results, using H@1 for measuring performance. 
Similarly to the experiments with source and target peers reported in Section~\ref{sec:rel_path_support_experiments}, we use incremental metrics, showing for each value of RPS the percentage and the H@1 of all the facts with support up to that value.

We observe that, for almost all models, greater RPS values lead to better performance. This proves that such models, to a certain extent, are capable of benefiting from longer-range dependencies. 

This correlation is visible in all datasets. It is particularly evident in WN18, WN18RR and YAGO3-10, and to a slightly lesser extent in FB15k-237. 
We point out that in FB15k-237 and WN18RR a significant percentage of facts displays a very low path support (less than 0.1). This is likely due to the filtering process employed to generate these datasets: removing facts breaks paths in the graph, thus making relational patterns less frequently observable.

FB15k is the dataset in which the correlation between RPS and performances, albeit clearly observable, seems weakest; we see this as a consequence of the severe test leakage displayed by FB15k. 
As a matter of fact, we have found evidence suggesting that, in presence of many relations with same or inverse meaning, models tend to focus on shorter dependencies for predictions, ignoring longer relational paths. We show this by replicating our experiment using RPS values computed with relational paths of maximum lengths 1 and 2, instead of 3. We report the FB15k chart Figure~\ref{fig:fb15k_paths_max1max2}, and the other charts in Appendix~\ref{appendix:rps_max1max2}. In FB15k and WN18, well known for their test leakage, the correlation with performances becomes evidently stronger. In FB15k-237, WN18RR and YAGO3-10, on the contrary, it weakens, meaning that 3-step relational paths are actually associated with correct predictions in these datasets.

Test leakage in FB15k and WN18 is actually so prominent that, on these datasets, we were able to use RPS as the scoring function of a standalone LP model based on observable features, obtaining acceptable results. 
We report the results of this experiment in Table~\ref{tab:rps_model}.
The evaluation pipeline is the same employed by all LP models and described in Section~\ref{sec:problem_overview}. 
Due to computational constraints, we use the RPS measure with paths up to 2 steps long, and use on each dataset a sample of 512 test facts instead of the whole test set (a single test fact can take more than 1h to predict). We do not run the experiment on YAGO3-10, on which the very high number of entities would make the ranking loop unfeasible. This experiment can be seen as analogous to the ones run by Toutanova~\emph{et~al}.~\cite{toutanova2015observed} and Dettmers~\emph{et~al}.~\cite{dettmers2018convolutional}, where simple models based on observable features are run on FB15k and WN18 to assess the consequences of their test leakage.

\begin{figure}
\centering
    \includegraphics[width=0.9\columnwidth]{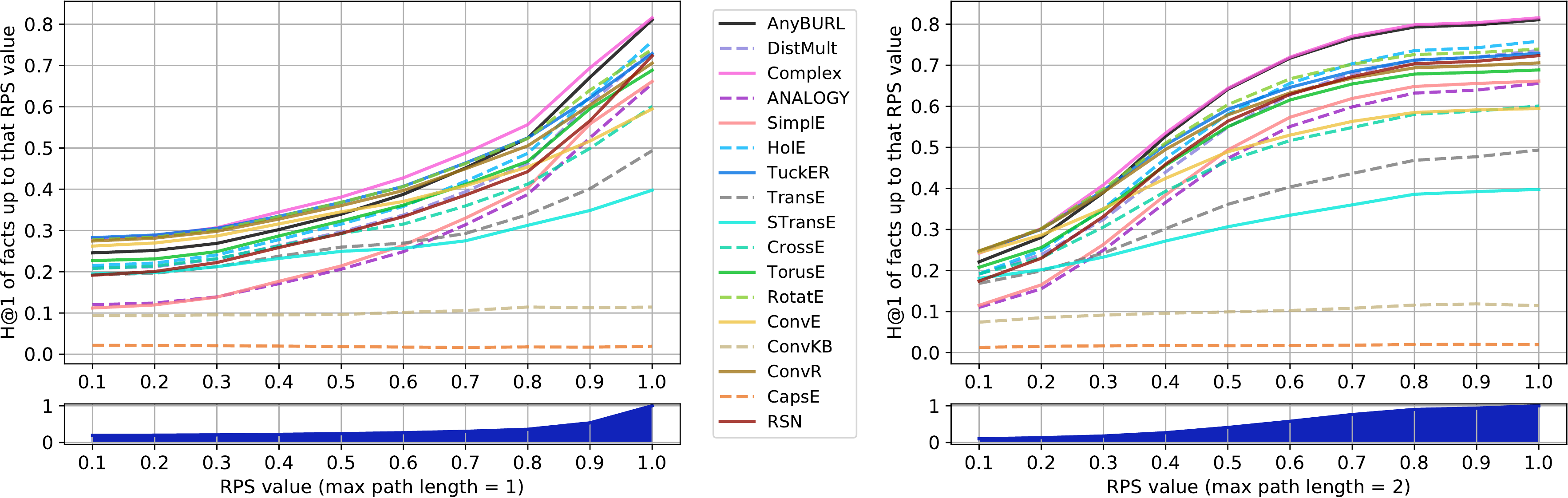}
    
    \caption{H@1 results for each LP model on FB15k varying the RPS of the test facts, computing RPS with paths up to length 1 and up to length 2.\label{fig:fb15k_paths_max1max2}}
\end{figure}

\begin{table}
\centering
    \includegraphics[width=\columnwidth]{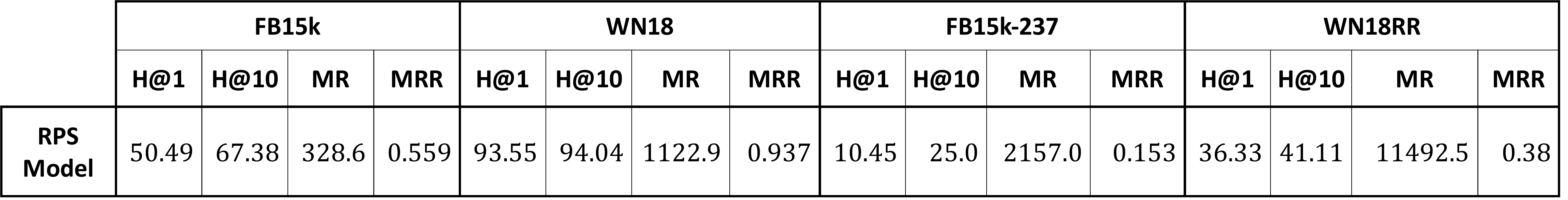}
    
    \caption{performances of an LP model based on observable features, using as a scoring function the RPS measure with relational paths up to 2 steps long. \label{tab:rps_model}}
\end{table}

\subsubsection{Relation Properties}
\label{sec:relprops_experiments}

\begin{figure}
     \centering
     \subfloat[FB15k\label{fig:fb15k_relprop}]{\includegraphics[width=0.3\textwidth]{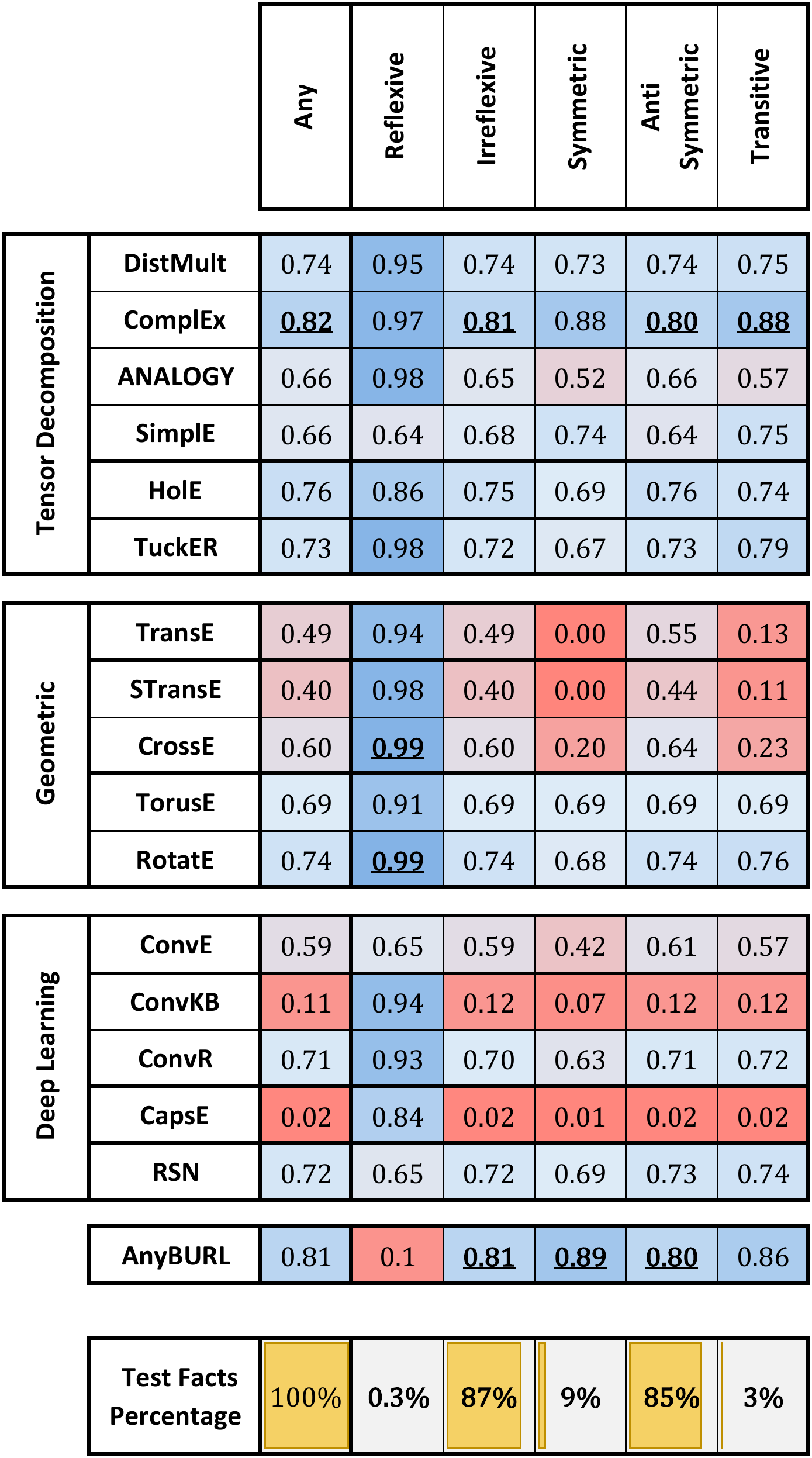}}
\hspace{1cm}
     \subfloat[FB15k-237\label{fig:fb15k237_relprop}]{\includegraphics[width=0.30\textwidth]{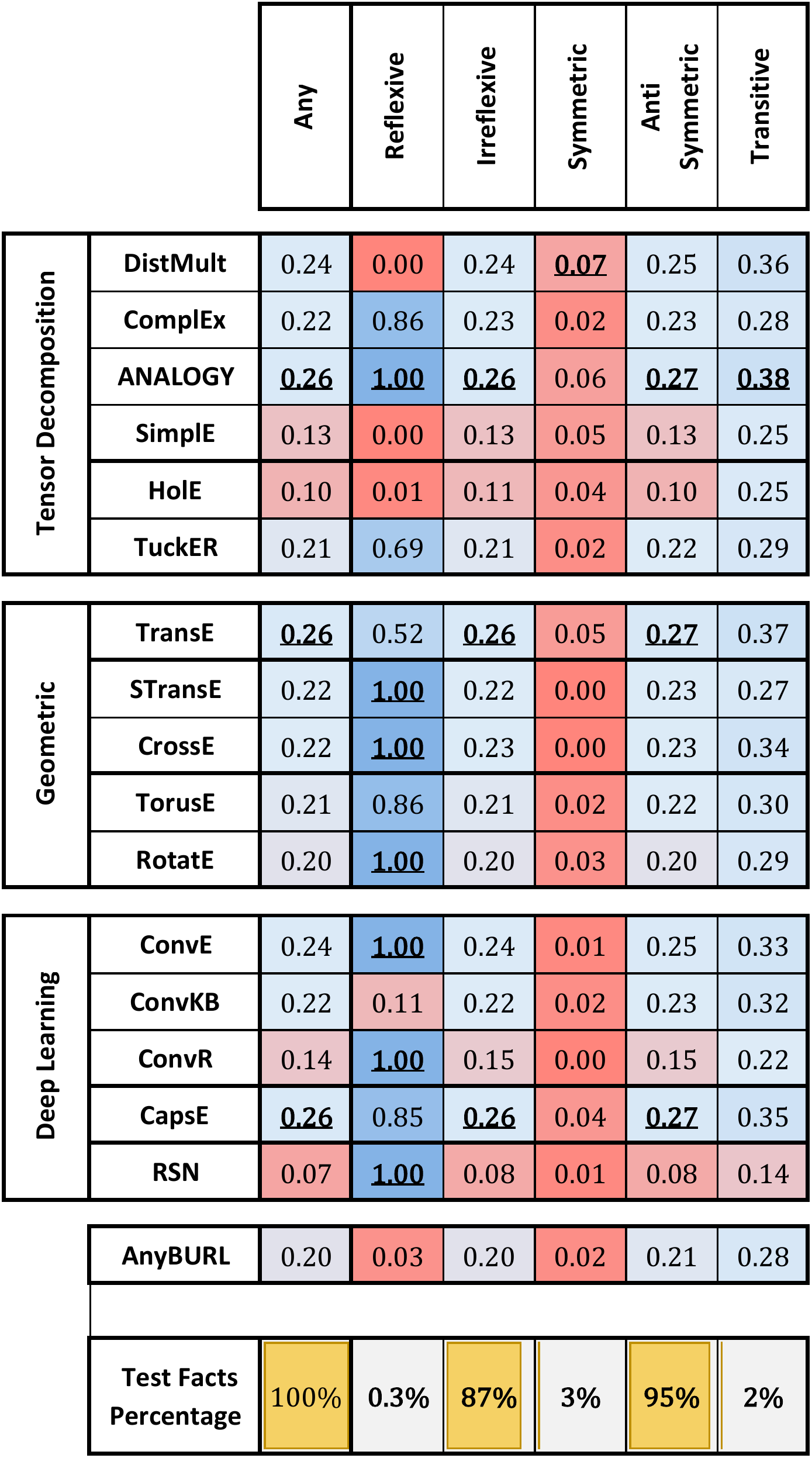}}
     
     \caption{H@1 results for each LP model on the \textbf{Freebase} datasets and corresponding percentages of test facts, for various relation properties. The best results for each column are in bold and underlined.}
\end{figure}

\begin{figure}
     \centering
     \subfloat[WN18\label{fig:wn18_relprop}]{\includegraphics[width=0.30\textwidth]{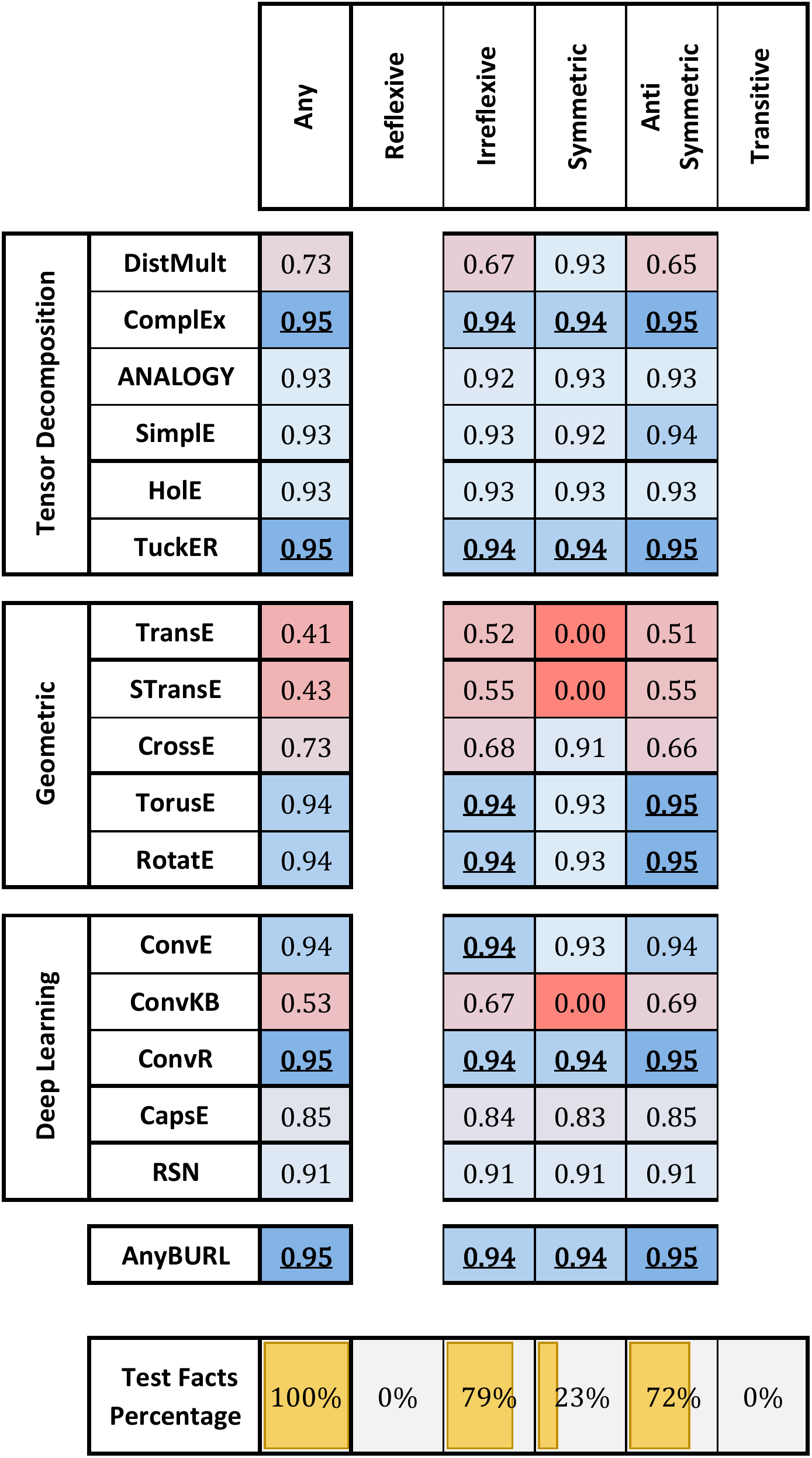}}
\hspace{1cm}
     \subfloat[WN18RR\label{fig:wn18rr_relprop}]{\includegraphics[width=0.3\textwidth]{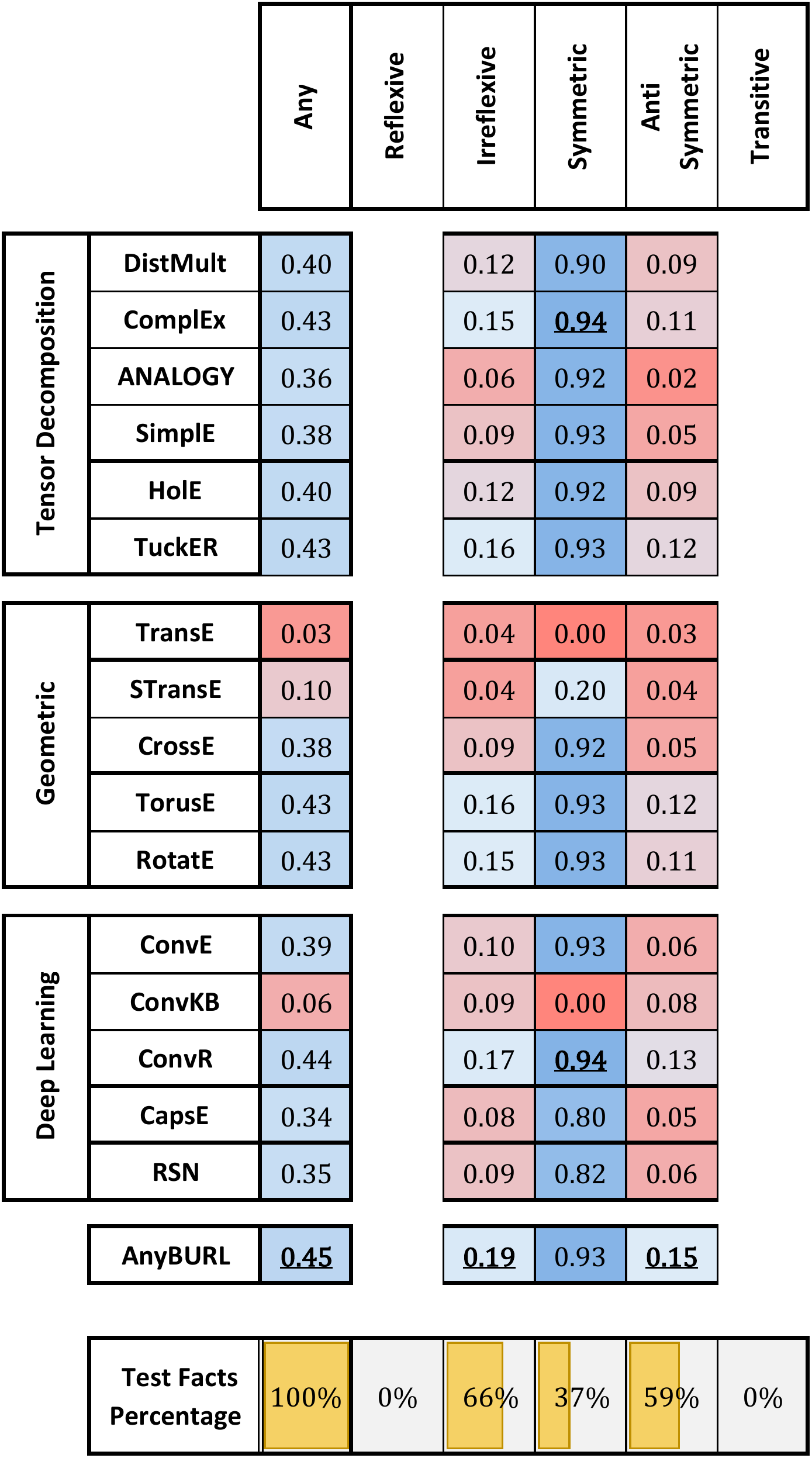}}
     \caption{H@1 results for each LP model on the \textbf{Wordnet} datasets, and corresponding percentages of test facts, for various relation properties. The best results for each column are in bold and underlined.}
\end{figure}

\begin{figure}
    \centering
    \subfloat[YAGO3-10]{\includegraphics[width=0.3\textwidth]{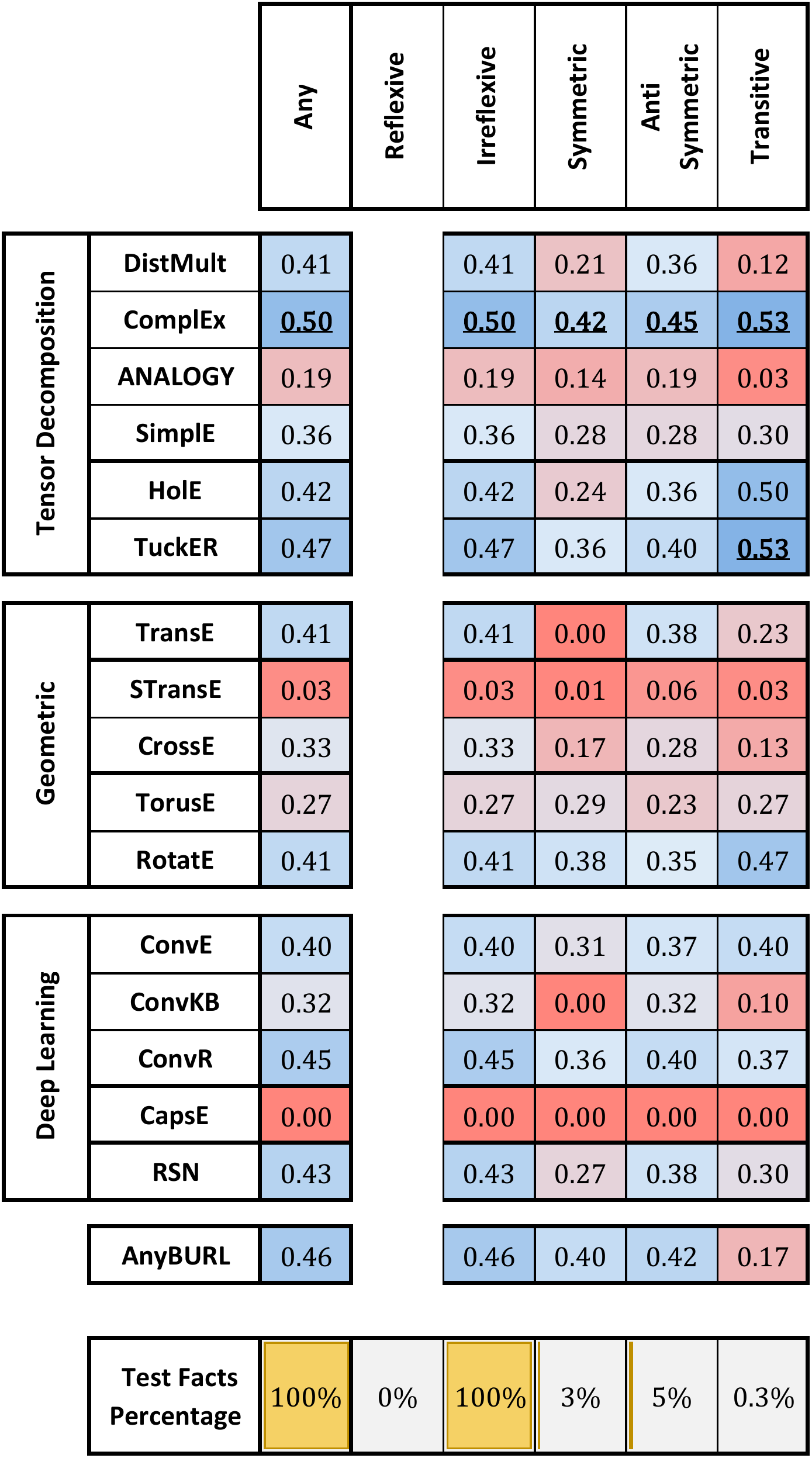}}
    \caption{H@1 results for each LP model on YAGO3-10, and corresponding percentages of test facts, for various relation properties. The best results for each column are in bold and underlined.\label{fig:yago_relprop}}
\end{figure}

Our goal in this experiment is to analyze how models perform when the relation in the fact to predict have specific properties. We take into account Reflexivity, Symmetry, Transitivity, Irreflexivity and Anti-symmetry, as already described in Section~\ref{sec:methodology:relprop}. 
We report in Figures~\ref{fig:fb15k_relprop}, \ref{fig:fb15k237_relprop}, \ref{fig:wn18_relprop}, \ref{fig:wn18rr_relprop}, \ref{fig:yago_relprop} our results, using H@1 for measuring effectiveness.

We divide test facts into buckets based on the properties of their relations. When a relation possesses multiple properties, the corresponding test facts are put in all the corresponding buckets. 
In all charts we include an initial bucket named \textit{any} containing all the test facts. For each model, this works as a global baseline, as it allows to compare the H@1 of each bucket to the global H@1 of that model.
Analogously to the distribution graphs observed in the previous experiments, for each bucket in each dataset we also report the percentage of test facts it contains.

In FB15K and FB15K-237 we observe an impressive majority of irreflexive and anti-symmetric relations. Only a few facts involve reflexive, symmetric or transitive relations. 
In WN18 and WN18RR the percentage of facts with symmetric relations is quite higher, but no reflexive and transitive relations are found at all. 
In YAGO3-10 all test facts feature irreflexive relations; there is a high percentage of facts featuring anti-symmetric relations as well, whereas only a few of them involve symmetric or transitive relations.

In FB15K all models based on embeddings seem to perform quite well on reflexive relations; on the other hand, the baseline AnyBURL~\cite{anyburl2019} obtains quite bad results on them possibly due to its rule-based approach We also observe that translational models such as TransE~\cite{transe}, CrossE~\cite{crosse} and STransE~\cite{stranse} struggle to handle symmetric and transitive relations, with very poor results. This problem seems alleviated by the introducton of rotational operations in TorusE~\cite{toruse} and RotatE~\cite{rotate}.

In FB15K-237, all models display globally worse performance; nonetheless, interestingly most of them manage to keep good performance on reflexive relations, the exceptions being ANALOGY~, SimplE~\cite{simple}, ConvE~\cite{dettmers2018convolutional} and RSN~\cite{rsn}. On the contrary they all display terrible performance in symmetric relations. This may depend on the sampling policy, that involves removing training facts connecting two entities when they are already linked in the test set: given any test fact \textlangle{}$h$,~$r$,~$t$\textrangle{}, even when $r$ is symmetric models can never see in training \textlangle{}$t$,~$r$,~$h$\textrangle{}.

In WN18 and WN18RR we observe a rather different situation.
This time, symmetric relations are easily handled by most models, with the notable exceptions of TransE~\cite{transe} and ConvKB~\cite{convkb}.
On WN18RR, the good results on symmetric relations balance, for most models, sub-par performance on irreflexive and anti-symmetric relations. 

In YAGO3-10 we observe once again TransE~\cite{transe} and ConvKB~\cite{convkb} having a hard time handling symmetric relations; on these relations, most models actually tend to behave a little worse than their global H@1.

\subsection{Reified Relation Degree}
\label{sec:reified_experiments}

Our goal in this experiment is to analyze how, in FreeBase-derived datasets, the degrees of the original reified relations affect predictive performance. Due to the properties of the S2C operations employed to explode reified relations into cliques, a higher degree of the original reified relation corresponds to a locally richer area in the dataset; therefore we expect such higher degrees to correspond to better performance.

We divide test facts into disjoint buckets based on the degree of the original reified relation, extracted as reported in Section~\ref{sec:methodology}. 

\begin{figure}
    \centering
    \subfloat[FB15k\label{fig:fb15k_reified_h1} ]{\includegraphics[width=.3\textwidth]{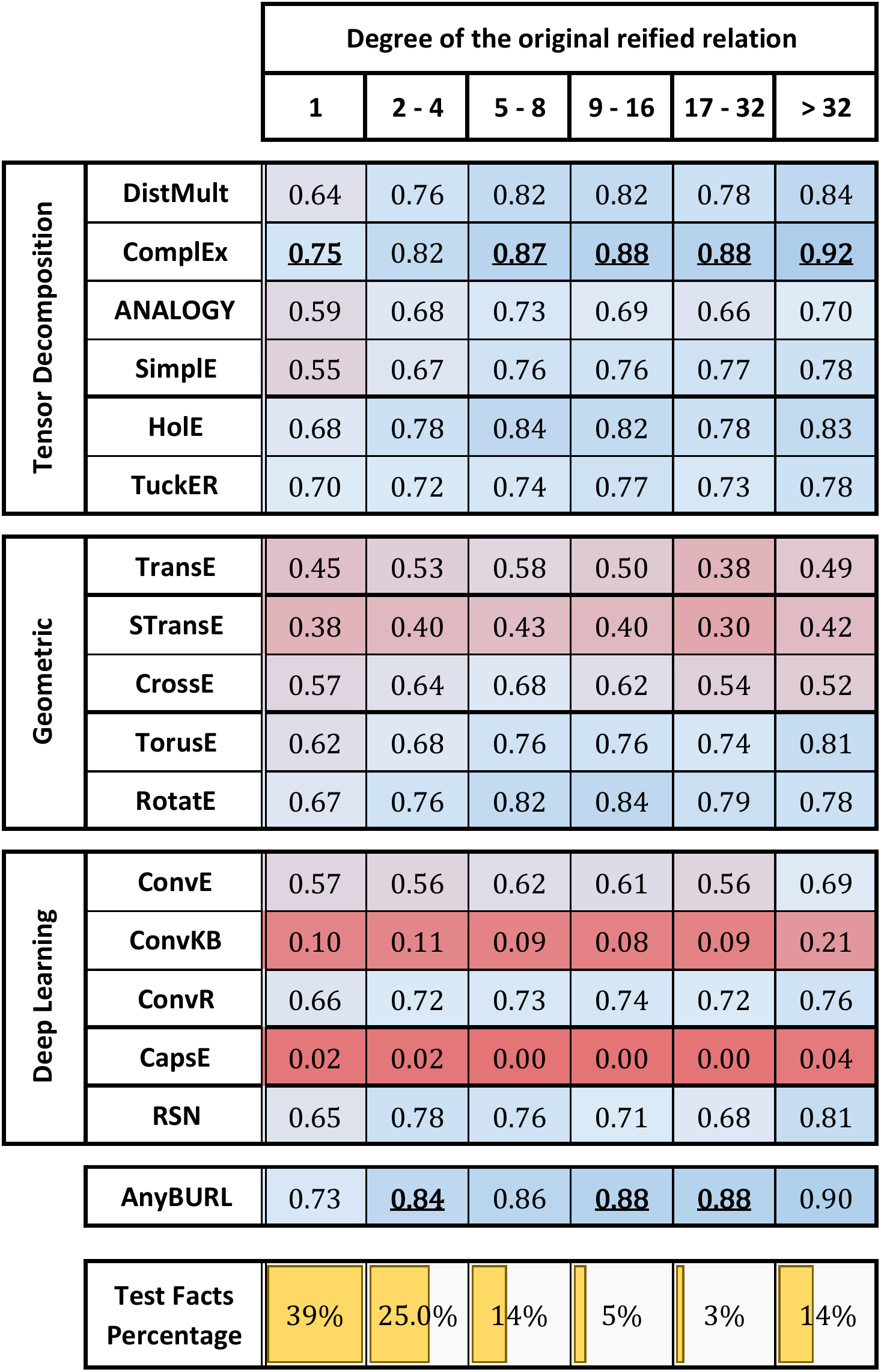}}
\hspace{1cm}
    \subfloat[FB15k-237\label{fig:fb15k237_reified_h1}]{\includegraphics[width=.3\textwidth]{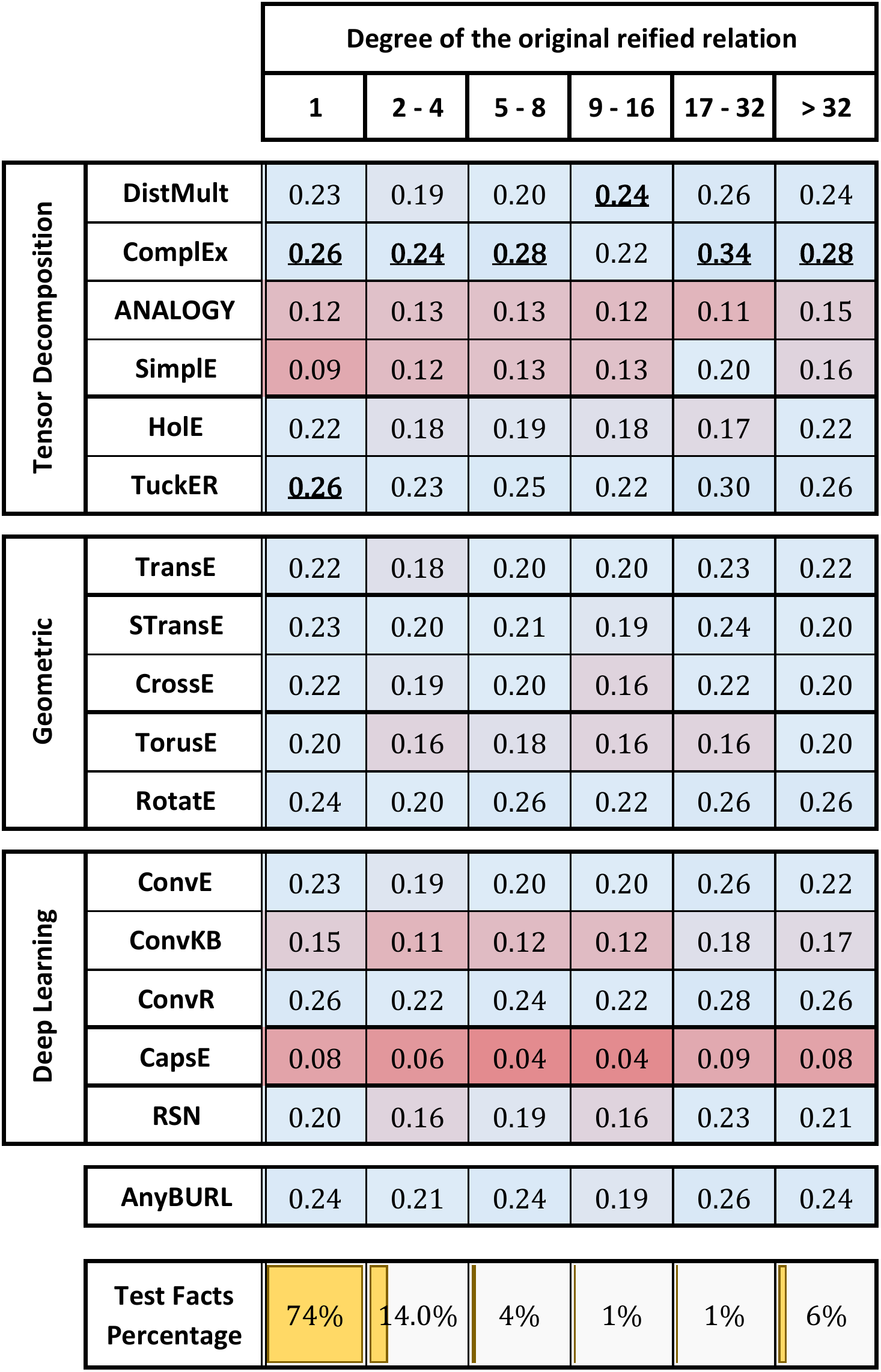}}
\caption{H@1 results for each LP model on the \textbf{Freebase} datasets, and corresponding distribution of test facts, varying the degree of the original reified relation in FreeBase. The best results for each column are marked in bold and underlined.}
\end{figure}

We compute the predictive performance of these buckets separately; we also include a separate bucket with degree value 1, containing the test facts that were not originated from reified relations in FreeBase. We report predictive performances using H@1 in Figures~\ref{fig:fb15k_reified_h1} and \ref{fig:fb15k237_reified_h1}. We also show, for each bucket, the percentage of test facts it contains with respect to the whole test set.

In FB15K, in most models we observe that a higher degree generally corresponds to better H@1. The main exceptions are TransE~\cite{transe}, CrossE~\cite{crosse} and STransE~\cite{stranse}, that show a stable or even worsening pattern. We found that, considering more permissive H@K metrics (e.g. H@10), all models, including these three, improve their performance; we explain this by considering that, due to the very nature of the S2C transformation, original reified relations tend to generate a high number of facts containing symmetric relations. TransE, STransE and CrossE are naturally inclined to represent symmetric relations with very small vectors in the embedding space: as a consequence, when learning facts with symmetric relations, these models tend to place the possible answers very close to each other in the embedding space. The result would be a crowded area in which the the correct target is often outranked when it comes to H@1, but manages to make it to the top K answers for larger values of K.

In FB15k-237 most of the redundant facts obtained from reified relations have been filtered away, therefore the large majority of test facts belongs to the first bucket. 

\subsubsection{Sensitivity to tie policy}
\label{sec:experiments:tie_policy}
We have observed that a few models, in their evaluation, are strikingly sensitive to the policy used for handling ties. This happens when models give the same score to multiple different entities in the same prediction: in this case results obtained with different policies diverge, and they are not comparable to one another anymore. In the most extreme case, if a model always gives the same score to all entities in the dataset, using \emph{min} policy it will obtain H@1 = 1.0 (perfect score) whereas using any other policy it would obtain H@1 around 0.0.

In our experiments we have found that CrossE~\cite{crosse} and, to a much greater extent, ConvKB~\cite{convkb} and CapsE~\cite{capse}, seem sensitive to this issue. Note that in their original implementations ConvKB and CapsE use \textit{min} policy by default, whereas CrossE uses \textit{ordinal} policy by default.

\begin{table}
    \includegraphics[width=\textwidth]{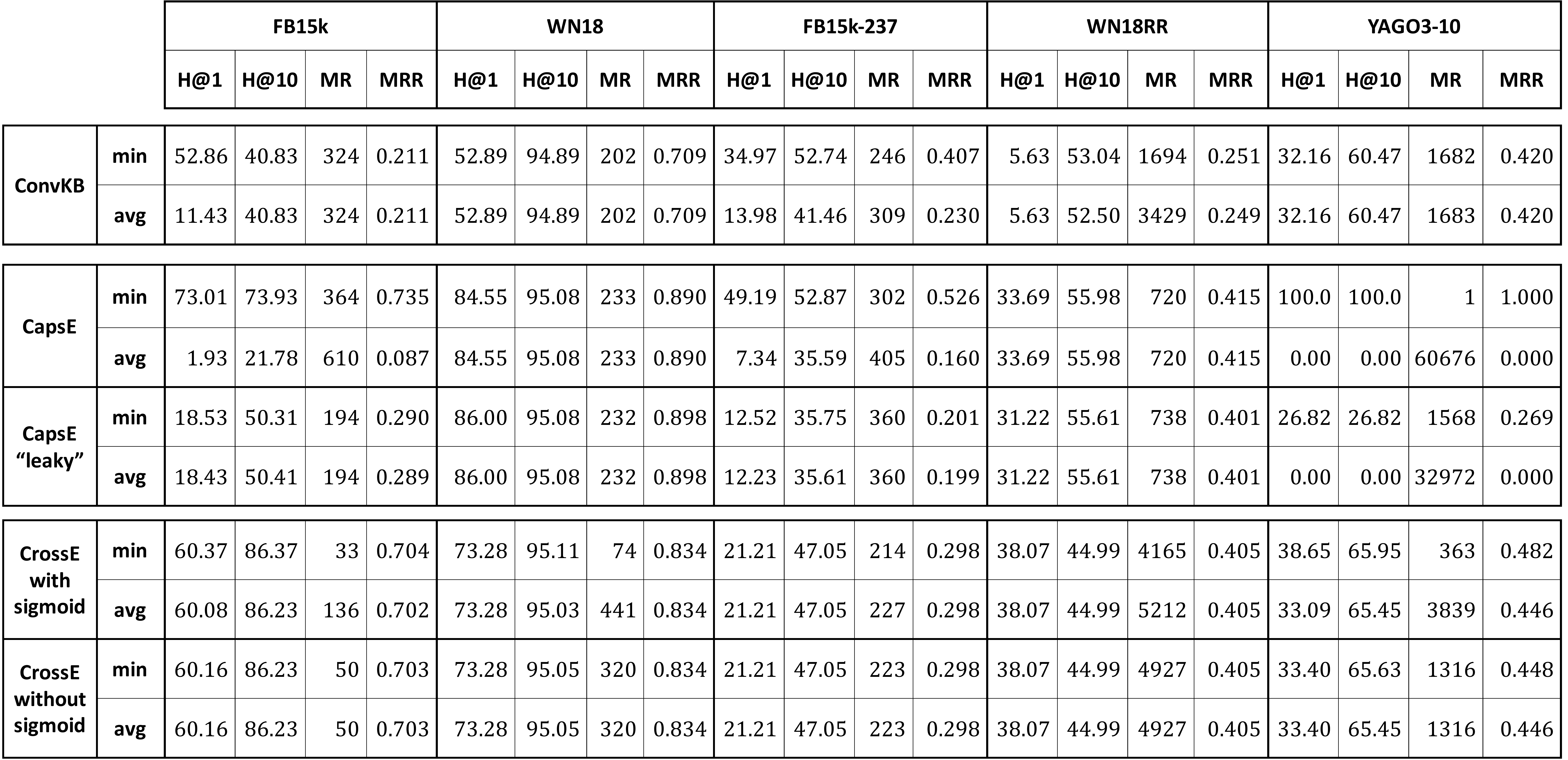}
    \caption{\label{tab:min_vs_avg} Results obtained with \textit{average} or \textit{ordinal} tie policy (\textbf{avg}) against results obtained with \textit{min} tie policy (\textbf{min}). The table features all the models for which these results show discrepancies (ConvKB; CapsE; CrossE), and the corresponding experiments (CapsE ``Leaky''; CrossE without sigmoid).}
\end{table}

In FB15k and FB15k-237 both ConvKB and CapsE display huge discrepancies on all metrics, whereas on WN18 and WN18RR the results are almost identical. On these datasets, no remarkable differences are observable for CrossE, except for MR, that is inherently sensitive to small variations. On YAGO3-10, quite interestingly, ConvKB does not seem to suffer from this issue, while CrossE shows a noticeable difference. CapsE shows the largest problems, with a behaviour akin to the extreme example described above.

We have run experiments on the architecture of these models in order to investigate which components are most responsible for these behaviours. We have found strong experimental evidence that \textit{saturating activation functions} may be the one of the main causes of this issue.

Saturating activation functions yield the same result for inputs beyond (or below) a certain value. For instance, the ReLU function (Rectified Linear Unit), returns 0 for any input lesser or equal to 0. Intuitively, saturating activation functions make it more likely to set identical scores to different entities, thus causing the observed issue. 

ConvKB and CapsE both use ReLUs between their layers. In order to verify our hypothesis, we have trained a version of CapsE substituting its ReLUs with Leaky ReLUs. The Leaky ReLU function is a non-saturating alternative to ReLU: it keeps a linear behaviour even for inputs lesser than 0, with slope $\alpha$ between 0 and 1. In our experiment we used $\alpha = 0.2$. We report in Table~\ref{tab:min_vs_avg} also the result of this CapsE variation, that we dubbed CapsE ``Leaky''. As a matter of fact, for CapsE ``Leaky'' the differences between results obtained with \textit{min} and \textit{average} are much less prominent. In FB15k and FB15k-237, differences in H@1 and MRR either disappear or decrease of 2 or even 3 orders of magnitude, becoming barely observable. In WN18 and WN18RR, much like in the original CapsE, no differences are observable. In YAGO3-10 we still report a significant difference between \textit{min} and \textit{average} results, but it is much smaller than before.

CrossE does not employ explicitly saturating activation functions; nonetheless, after thoroughly investigating the model architecture, we have found that in this case too the issue is rooted in saturation. As shown in the scoring function in table~\ref{tab:loss_table}, CrossE normalizes its scores by applying a sigmoid function in a final step. The sigmoid function is not saturating, but for values with very large modulus its slope is almost nil: therefore, due to low-level approximations, it behaves just like a saturating function.
Therefore, we tested again CrossE just removing the sigmoid function in evaluation; since the sigmoid function is monotonous and growing, removing it does not affect entity ranks. In the obtained results all discrepancies between \textit{min} and \textit{average} policies disappear. As before, we report the results in Table~\ref{tab:min_vs_avg}.

For all the other models in our analysis we have not found significant differences among their results obtained with different policies.

\section{Key Takeaways and Research Directions}
\label{sec:lessons}
In this section we summarize the key takeaways from our comparative analysis. We believe these lessons can inform and inspire future research on LP models based on KG embeddings.

\subsection{Effect of the design choices}

We discuss here comprehensive observations regarding the performances of models as well as their robustness across evaluation datasets and metrics. We report findings regarding trends investing entire families based on specific design choices, as well as unique feats displayed by individual models.

Among those included in our analysis, Tensor Decomposition models show the most solid results across datasets. 
In the implementations taken into account, most of these systems display uniform performances on all evaluation metrics across the datasets of our analysis (with the potential exceptions of ANALOGY and SimplE, that are seemingly more fluctuating).
In particular, ComplEx with its N3 regularization displays amazing results on all metrics across all datasets, being the only embedding-based model consistently comparable to the baseline AnyBURL.

The Geometric family, on the other hand, shows slightly more unstable results. In the past years, research has devoted a considerable effort into translational models, ranging from TransE to its many successors with multi-embedding policies for handling many-to-one, one-to-many and many-to-many relations. These models show interesting results, but still suffer from some irregularities across metrics and datasets. 
For instance, models such as TransE and STransE seem to particularly struggle on the WN18RR dataset, especially when it comes to H@1 and MRR metrics. 
All in all, models relying solely on translations seem to have been outclassed by recent roto-translational ones. 
At this regard, RotatE shows remarkably consistent performances across all datasets, and it particularly shines when taking into account H@10.

Deep Learning models, finally, are the most diverse family, with wildly different results depending on the architectural choices of the models and on their implementations. 
ConvR and RSN display by far the best results in this family, achieving very similar, state-of-the-art performance in FB15k, WN18 and YAGO3-10. In FB15k-237 and WN18RR, whose filtering processes have cut away the most relevant paths, RSN seems to have a harder time, probably due to its formulation that explicitly leverages paths. 
On the other hand, models such as ConvKB and CapsE often achieve promising results on H@10 and MR metrics, whereas they seem to struggle with H@1 and MRR; furthermore, in some datasets they are clearly hindered by their issues with tie policies described in Section~\ref{sec:experiments:tie_policy}.

We stress that in the LP task the rule-based AnyBURL proves to be a remarkably well-performing model, as it consistently ranks among the best models across almost all datasets and metrics.

\subsection{The importance of the graph structure}

We have shown consistent experimental evidence that graph structural features have a large influence on what models manage to learn and predict.

We observe that in almost all models and datasets, predictions seem to be facilitated by the presence of source peers and hindered by the presence of target peers. As already mentioned, source peers work as examples that allow models to characterize more effectively the relation and the target to predict, whereas target peers lead models to confusion, as they try to optimize embeddings to fit too many different answers for the same question.

We also observe evidence suggesting that almost all models -- even across those that only learn individual facts in training -- seem able to leverage to some extent relational paths and patterns.

All in all, the toughest scenarios for LP models seem to take place when there are relatively more target peers than source peers, in conjunction with a low support offered by relational paths. In these cases, models usually tend to show quite unsatisfactory performances. We believe that these are the areas where future research has most room for improvement, and thus the next big challenges to address in the LP research field.

We also point out interesting differences in behaviours and correlations depending on the features of the employed dataset.
In FB15k and WN18, which display strong test leakage, model performances show a prominent correlation with the support provided by shorter relational paths, with length 1 or 2. This is likely caused by such short paths including relations with inverse meaning or same meaning as the relation in the facts to predict. 
On the contrary, in their FB15k-237, WN18RR and YAGO3-10, which do not suffer from test leakage, models appear to rely also on longer relational paths (3 steps), as well as on the numbers of source/target peers. 

We believe that this leaves room for intriguing observations. 
In presence of very short patterns providing overwhelming predictive evidence (e.g., the inverse relations that cause test leakage), models seem very prone to just focusing on them, disregarding other forms of reasoning: this can be seen as an unhealthy consequence of the test leakage problem. In more balanced scenarios, on the contrary, models seem to investigate to a certain extent longer dependencies, as well as to focus more on analogical reasoning supported by examples (such as source peers). 

We also observe that applying LP models based on embeddings to infer relations with cardinality greater than 2 is still an open problem.
As already mentioned in Section~\ref{sec:methodology:reified}, the FreeBase KG represents hyperedges as reified CVT nodes.
Hyperedges constitute the large majority of edges in FreeBase: as noted by Fatemi~\emph{et~al}.~\cite{hype} and Wen~et~al.~\cite{wen2016binary}, 61\% of the FreeBase relations are beyond-binary, and the corresponding hyperedges involve more than 1/3rd of the FreeBase entities. The FB15k and FB15k-237 datasets have been built by performing S2C explosion on FreeBase subsamples; this has resulted in greatly altering both the graph structure and its semantics, with overall loss of information. We believe that, in order to assess the effects of this process, it would be fruitful to extract novel versions of FB15k and FB15k-237 in their original reified structure without applying S2C. We also note that models such as m-TransH~\cite{wen2016binary} and the recent HypE~\cite{hype} have tried to circumvent these issues by developing systems that can explicitly learn hyperedges. Despite them being technically usable on datasets with binary relations, of course their unique features emerge best when dealing with relations beyond binary.

\subsection{The importance of tie policies}

We report that differences in the policies used to handle score ties can lead to huge differences in predictive performance in evaluation. 
As a matter of fact, such policies are today treated as almost negligible implementation details, and they are hardly ever even reported when presenting novel LP models.
Nevertheless, we show that performances computed relying on different policies risk to be not directly comparable to one another, and might not even reflect the actual predictive effectiveness of models. 
Therefore we strongly advise researchers to use the same policy in the future; in our opinion, the ``average'' policy seems the most reasonable choice.
We have also found strong experimental evidence that saturating activation functions, such as ReLU, play a key role in leading models to assign the same scores to multiple entities in the same prediction; approximations may also lead non-saturating functions, such as Sigmoid, behave as saturating in regions where their slope is particularly close to 0.

\section{Related Works}
\label{sec:related}
Works related to ours can be roughly divided into two main categories: \emph{analyses} and \emph{surveys}.
Analyses usually run further experiments trying to convey deeper understandings on LP models, whereas surveys usually attempt to organize them into comprehensive taxonomies based on their features and capabilities.

\paragraph{Analyses}
Chandrahas~\emph{et~al.}~\cite{conicity} study geometrical properties of the obtained embeddings in the latent space. They separate models into additive and multiplicative and measure the \emph{Alignment To Mean} (ATM) and \emph{conicity} of the learned vectors, showing that additive models tend to learn significantly sparser vectors than multiplicative ones. They then check how this reflects on the model peformances. Their observations are intriguing, especially for multiplicative models, where a high conicity (and thus a low vector spread) seems to correlate to better effectiveness. 

Wang~\emph{et~al}.~\cite{mapAtK} provide a critique on the current benchmarking practices. They observe that current evaluation practices only compute the rankings for test facts; therefore, we are only verifying that, when a question is "meaningful" and has answers, our models prioritize the correct ones over the wrong ones. This amounts to performing question answering rather than KG completion, because we are not making sure that questions with no answers (and therefore not in the dataset) result in low scores. Therefore, they propose a novel evaluation pipeline, called Entity-Pair Ranking (PR) including all possible combinations in $\mathcal{E} \times \mathcal{R} \times \mathcal{E}$. We wholly agree with their observations; unfortunately, we found that for our experiments, where the full ranking for all predictions is required for all models in all datasets, PR evaluation is way too time-consuming and thus unfeasible.

Akrami~\emph{et~al}.\cite{akrami2018re} use the same intuition as Toutanova~\emph{et~al}.~\cite{toutanova2015observed} to carry out a slightly more structured analysis, as they use a wider variety of models to check the performance gap between FB15k and FB15K-237.

Kadlec~\emph{et~al}.~\cite{kadlec2017knowledge} demonstrate that a carefully tuned implementation of DistMult~\cite{distmult} can achieve state-of-the-art performances, surpassing most of its own successors, raising questions on whether we are developing better LP models or just tuning better hyperparameters.

Tran~\emph{et~al}.~\cite{2019multiEmbeddingInteractionPerspective} interpret 4 models based on matrix factorization as special cases of the same multi-embedding interaction mechanism. 
In their formulation, each KG element $k$ is expressed as a set of vectors $\{\boldsymbol{k}^{(1)}, \boldsymbol{k}^{(2)}, ..., \boldsymbol{k}^{(n)}\}$; the scoring functions combine such vectors using trilinear products. The authors also include empirical analyses and comparisons among said models, and introduce a new multi-embedding one based on quaternion algebra.

All the above mentioned analyses have a very different scope from ours. Their goal is generally to address specific issues or investigate vertical hypotheses; on the other hand, our objective is to run an extensive comparison of models belonging to vastly different families, investigating the effects of distinct design choices, discussing the effects of different benchmarking practices and underlining the importance of the graph structure.

\paragraph{Surveys}
Nickel~\emph{et~al.}~\cite{nickel2015review} provide an overview for the most popular techniques in the whole field of Statistic Relational Learning, to which LP belongs. The authors include both traditional approaches based on observable graph features and more recent ones based on latent features. Since the paper has been published, however, a great deal of further progress has been made in KG Embeddings.

Cai~\emph{et~al.}~\cite{cai2018comprehensive} provide a survey for the whole Graph Embedding field. Their scope is not limited to KGs: on the contrary, they overview models handling a wide variety of graphs (Homogeneous, Heterogeneous, with Auxiliary Information, Constructed from Non-Relational Data) with an even wider variety of techniques. Some KG embedding models are briefly discussed in a section dedicated to models that minimize margin-based ranking loss. 

To this end, the surveys by Wang~\emph{et~al.}~\cite{survey_zhendong} and by Nguyen~\cite{survey_nguyen} are the most relevant to our work, as they specifically focus on KG Embedding methods. 
In the work by Wang~\emph{et~al.}~\cite{survey_zhendong}, models are first coarsely grouped based on the input data they rely on (facts only; relation paths; textual contents; etc); the resulting groups undergo further finer-grained selection, taking into account for instance the nature of their scoring functions (e.g. distance-based or semantic-matching-based). What's more, they offer detailed descriptions for each of the models they encompass, explicitly stating its architectural peculiarities as well as its space and time complexities. Finally, they take into account a large variety of applications that the Knowledge Graph Embedding models can support.
The work by Nguyen~\cite{survey_nguyen} is similar, albeit more concise, and also includes current state-of-the-art methods such as RotatE~\cite{rotate}.

Our work is fundamentally different from these surveys: while they only report results available in the original papers, we design experiments to extensively investigate the empirical behaviours of models. As discussed in Section~\ref{sec:intro}, results reported in the original papers are generally obtained in very different settings and they are generally global metrics on the whole test sets; as a consequence, it is difficult to interpret and compare them.

\section{Conclusions}
\label{sec:concl}
In this work we have presented the first extensive comparative analysis on LP models based on KG embedding.

We have surveyed 16 LP models representative of diverse techniques and architectures, and we have analyzed their efficiency and effectiveness on the 5 most popular datasets in literature.

We have introduced a set of structural properties characterizing the training data, and we have shown strong experimental evidence that they produce paramount effects on prediction performances. In doing so, we have investigated the circumstances that allow models to perform satisfactorily, while identifying the areas where research still has room for improvement.

We have thoroughly discussed the current evaluation practices, verifying that they can rely on different low-level policies producing incomparable and, in some cases, misleading results. We have analyzed the components that make models most sensitive to these policies, providing useful observations for future research. 

\section*{Acknowledgments} 
We thank proff. Alessandro Micarelli and Fabio Gasparetti for providing the computational resources employed in this research. We thank Simone Scardapane, Matteo Cannaviccio and Alessandro Temperoni for their insightful discussions.
We heartfully thank the authors of AnyBURL, ComplEx-N3, ConvR, CrossE and RSN for their for their amazing support and guidance on their models. 

\bibliographystyle{abbrv}
\bibliography{references}

\clearpage
\appendix
\section{Hyperparameters}
\label{appendix:hyperparams}

We report here the hyperparameter setting used for each model in our experiments. We highlight in yellow the settings we have found manually, and report in the $Space$ column the size of the corresponding space of combinations.
\begin{table}[H]
    \centering{
        \includegraphics[width=0.82\textwidth]{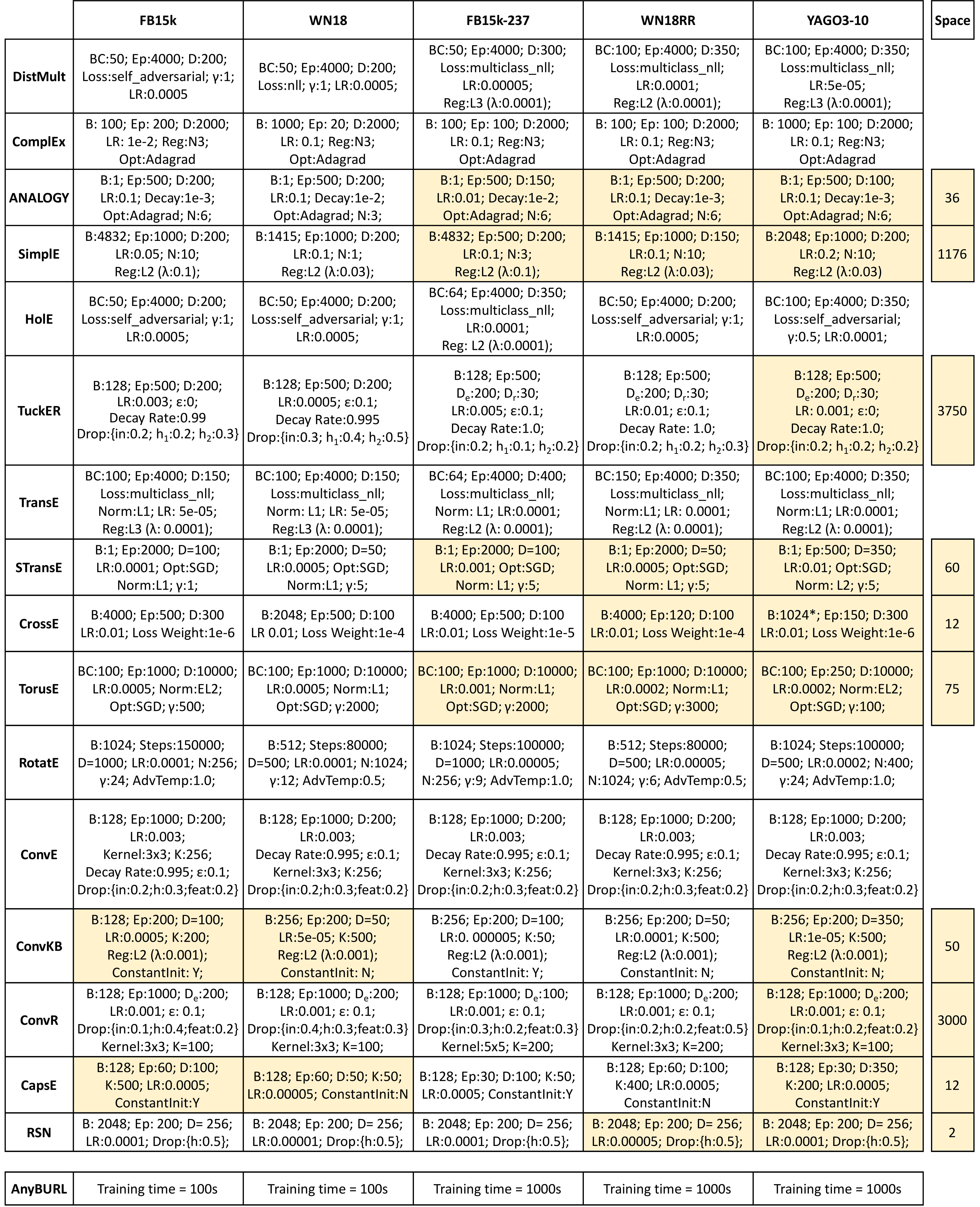}
        \caption{\label{tab:hyperparams_table} Hyperparameters used to train all the models in our work. $B$: batch size; alternatively $BC$: batch count. 
        $Ep$: training epochs; alternatively $Steps$: training steps. 
        $D$: embedding dimension; alternatively, $D_e$ and $D_r$: entity and relation embedding dimension. 
        $LR$: learning rate. $\gamma$: regularization margin. 
        $Reg$: regularization method; $\lambda$: lambda for $Reg \in \{L1, L2, L3\}$. 
        $\epsilon$: label smoothing.
        $Opt$: optimizer (default: $ADAM$). 
        $K$: convolutional filters. $Kernel$: convolutional kernel.
        $Drop$: dropout rate ($in$: in input; $h_i$: in $i$-th hidden layer; $feat$: in features). 
        $N$: negative samples per training fact (default: 1).
        $AdvTemp$: temperature in adversarial negative sampling.
        $ConstantInit$: initialize filters as [0.1, 0.1, -0.1] if $Y$, otherwise from a truncated normal distribution.}
    }
\end{table}

\clearpage
\appendix
\section{RPS with paths of maximum lengths 1 and 2}
\label{appendix:rps_max1max2}

\begin{figure}[h]
    \subfloat[FB15k-237\label{fig:fb15k-237_rps_max1max2}]
    {\includegraphics[width=0.9\columnwidth]{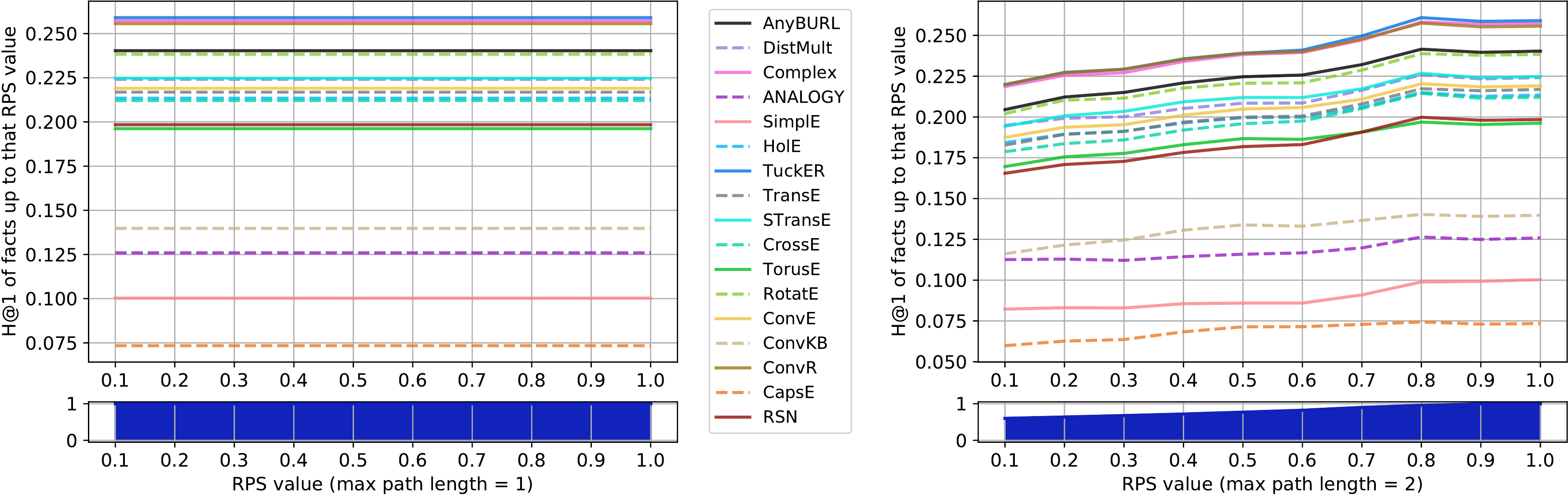}}
    \caption{H@1 results for each LP model on FB15k-237 varying the RPS of the test facts, computing RPS with paths up to length 1 and up to length 2.}
\end{figure}

\begin{figure}[h]
    \subfloat[WN18\label{fig:wn18_rps_max1max2}]{\includegraphics[width=0.9\columnwidth]{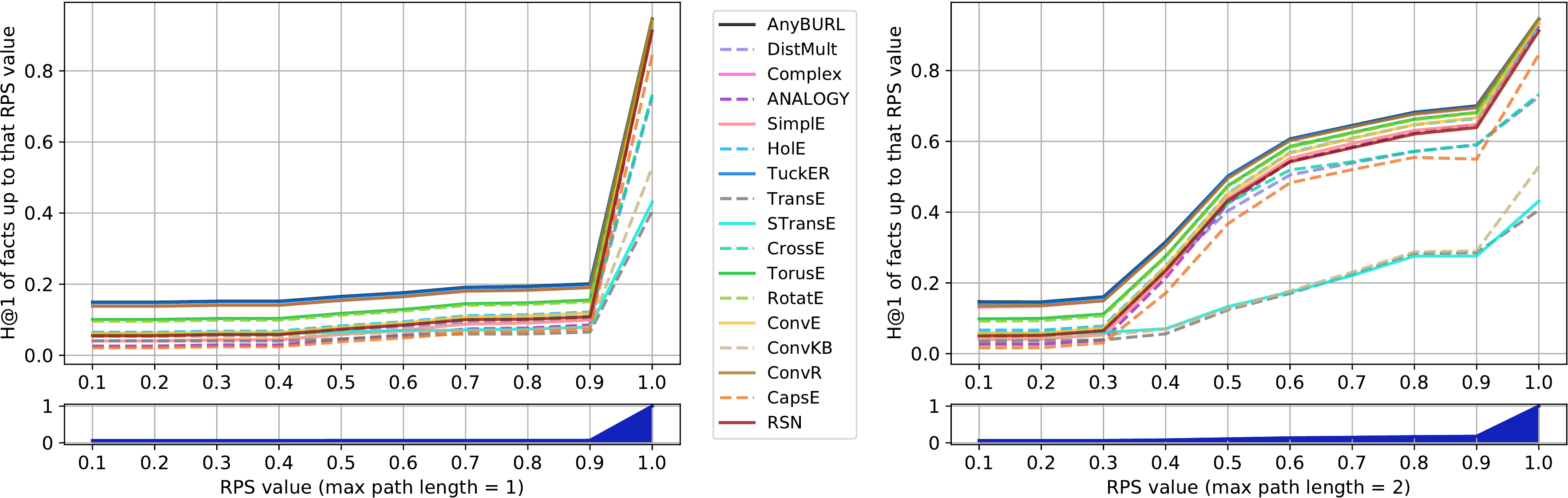}}

    \subfloat[WN18RR\label{fig:wn18rr_rps_max1max2}]{\includegraphics[width=0.9\columnwidth]{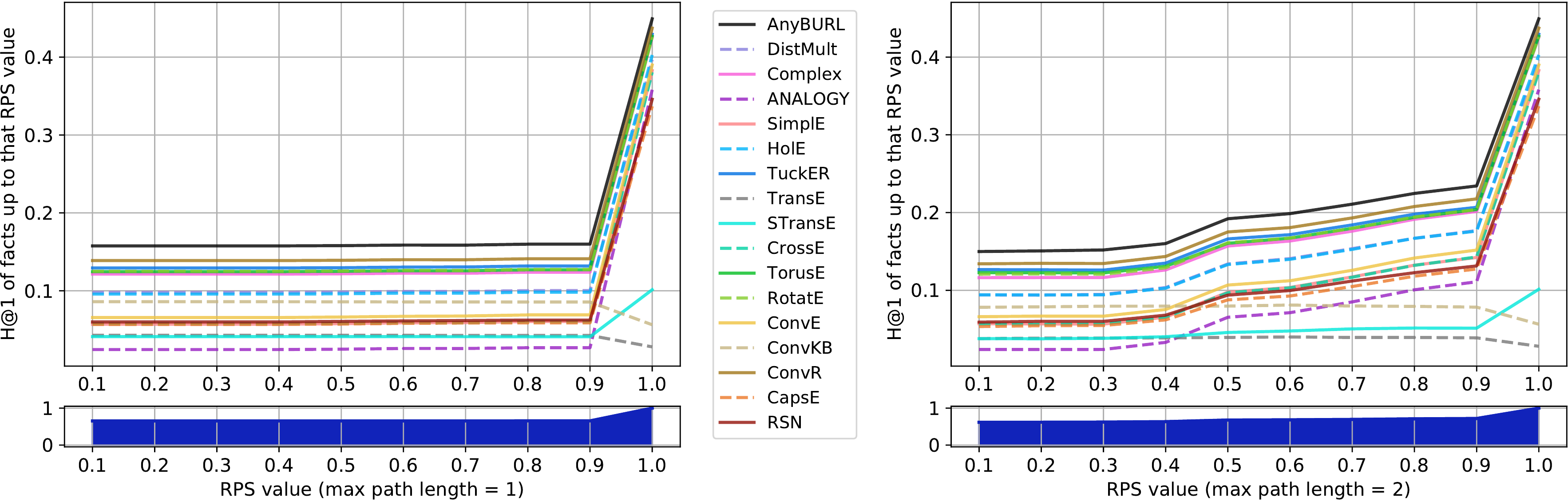}}
    \caption{H@1 results for each LP model on WordNet datasets varying the RPS of the test facts, computing RPS with paths up to length 1 and up to length 2.}
\end{figure}

\begin{figure}[h]
    \subfloat[YAGO3-10\label{fig:yago310_rps_max1max2}]{\includegraphics[width=0.9\columnwidth]{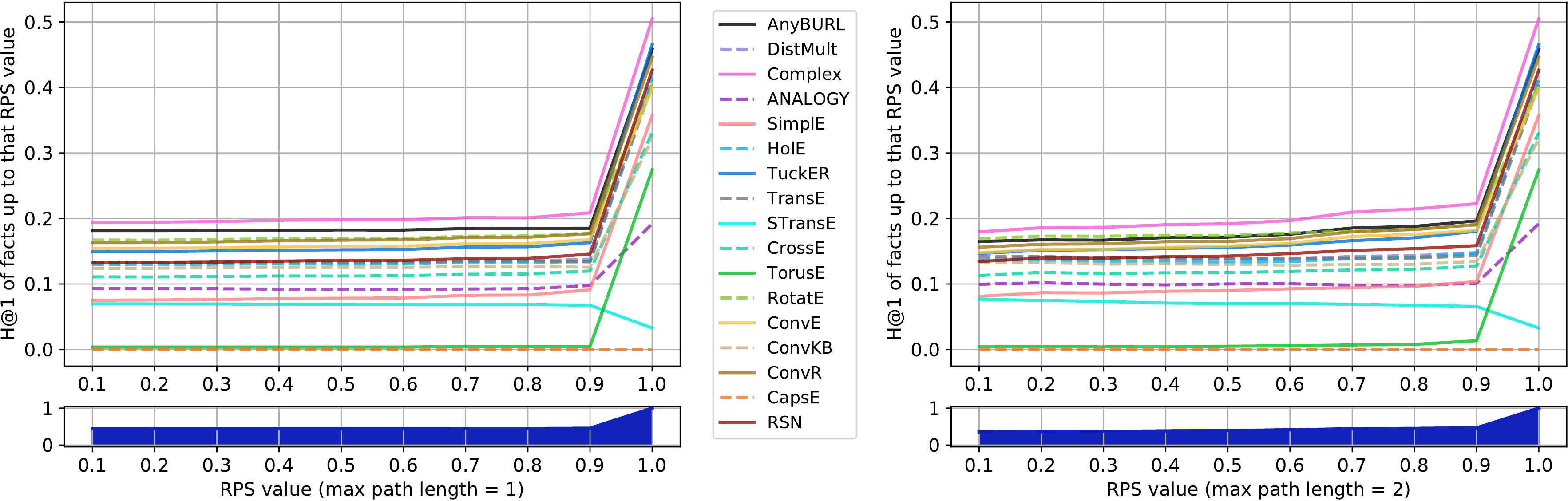}}
    \caption{H@1 results for each LP model on Yago datasets varying the RPS of the test facts, computing RPS with paths up to length 1 and up to length 2.}
\end{figure}

\end{document}